\PassOptionsToPackage{table}{xcolor}  % the class loads xcolor; the tables use \rowcolor / \cellcolor
\documentclass[11pt, a4paper]{google}  % no copyright option: arXiv 2605.14389 has no copyright line
\DisableLigatures{encoding = T1, family = tt*}
\renewcommand{\titlefont}{\color{black}\normalfont\bfseries\fontsize{20}{23}\selectfont}
\pdftrailerid{redacted}

\usepackage{natbib}
\usepackage{subcaption}
\usepackage{amsfonts}       % blackboard math symbols
\usepackage{nicefrac}       % compact symbols for 1/2, etc.
\usepackage{tabularray}
\usepackage{svg}            % Figure 2; see the note at the top about svg-inkscape/
\usepackage{xspace}
\usepackage{float}
\usepackage{wrapfig}        % Tables 2 and 3 stand at the right of the text that runs beside them
\usepackage{multirow}
\usepackage{algorithm}             % for the algorithm environment
\usepackage[noend]{algpseudocode}  % for pseudocode formatting
\usepackage[most]{tcolorbox}       % for the prompt boxes
\usepackage{afterpage}
\tcbset{lines before break=4, before upper={\frenchspacing}}
\usepackage{listings}              % for the raw text inside the boxes
\usepackage{xurl}

\mathchardef\UrlBigBreakPenalty=700
\mathchardef\UrlBreakPenalty=1000
\hypersetup{colorlinks=true, citecolor=blue, linkcolor=cyan, urlcolor=green}  % as in arXiv 2605.14389
\hypersetup{pdftitle={Nexus: An Agentic Framework for Context-Aware Time Series Forecasting},
  pdfauthor={Sarkar Snigdha Sarathi Das; Palash Goyal; Mihir Parmar; Nanyun Peng; Vishy Tirumalashetty; Chun-Liang Li; Rui Zhang; Jinsung Yoon; Tomas Pfister}}

\usepackage{amsmath,amsfonts,bm}

\def\eqref#1{equation~\ref{#1}}
\def\1{\bm{1}}

\DeclareMathAlphabet{\mathsfit}{\encodingdefault}{\sfdefault}{m}{sl}
\SetMathAlphabet{\mathsfit}{bold}{\encodingdefault}{\sfdefault}{bx}{n}

\newcommand{\ours}{\textsc{Nexus}}
\newcommand{\leaf}{\textsc{Leaf}}

\renewcommand{\algorithmicrequire}{\textbf{Input:}}
\renewcommand{\algorithmicensure}{\textbf{Output:}}

\UseMicrotypeSet[protrusion]{alltext-nott}
\newlength{\nexusintextsep}
\uselogo{}

\title{\ours{}: An Agentic Framework for Context-Aware Time Series Forecasting}

\correspondingauthor{ \{sarkardas, palashgoyal,mihirparmar\}@google.com}

\renewcommand{\today}{}

\makeatletter
\let\ps@nexusfirstbase\ps@firststyle
\def\ps@firststyle{\ps@nexusfirstbase
  \fancyfoot[L]{\footerfont $^\clubsuit$Please correspond with \the\correspondingauthor}}
\makeatother

\author[1]{\fontsize{10.0pt}{10.0pt}\selectfont Sarkar Snigdha Sarathi Das}
\author[1]{\fontsize{10.0pt}{10.0pt}\selectfont Palash Goyal}
\author[1]{\fontsize{10.0pt}{10.0pt}\selectfont Mihir Parmar}
\author[1]{\fontsize{10.0pt}{10.0pt}\selectfont Nanyun Peng}
\author[1]{\fontsize{10.0pt}{10.0pt}\selectfont Vishy Tirumalashetty}
\author[1]{\fontsize{10.0pt}{10.0pt}\selectfont Chun-Liang Li}
\author[2]{\fontsize{10.0pt}{10.0pt}\selectfont Rui Zhang}
\author[1]{\fontsize{10.0pt}{10.0pt}\selectfont Jinsung Yoon}
\author[1]{\fontsize{10.0pt}{10.0pt}\selectfont Tomas Pfister}

\affil[1]{\fontsize{9.0pt}{9.0pt}\selectfont Google}
\affil[2]{\fontsize{9.0pt}{9.0pt}\selectfont Pennsylvania State University}

\begin{abstract}
Language models have emerged as highly capable reasoners, yet their performance in time series forecasting remains uneven across domains and levels of contextual
grounding. This is primarily due to their suboptimal time series reasoning and numerical modeling capabilities. To tackle this, we introduce
\ours{}, a multimodal forecasting framework that bridges qualitative event reasoning with structural numerical modeling through principled statistical synthesis. \ours{} follows three structured stages: (1) Temporal profiling and anchor generation, which distills the multimodal history into a timestamped causal-temporal profile and generates a forecast anchor (2) Dual contextual outlook generation, which produces decoupled macro and micro level forecast trajectories, and (3) Bayesian synthesis, which reconciles the forecasting anchor with both contextual forecast trajectories via closed-form Gaussian updating weighted by empirical error covariances. Evaluated across three diverse forecasting benchmarks constituting eight datasets spanning rich event contexts, abrupt shocks, and purely numerical series, \ours{} consistently outperforms leading time-series foundation models and language model baselines, while providing interpretable causal reasoning.

\end{abstract}

\begin{document}

\maketitle
\suppressfloats[t]  % no float above the title block: Figure 1 goes to the top of page 2, as in arXiv 2605.14389

\providecommand{\todo}[1]{\textcolor{red}{\textbf{[TODO: #1]}}}

\section{Introduction}
\label{sec:introduction}

% Teaser: caption beside the image; \suppressfloats[t] in main.tex sends it to the top of page 2.
% \begin{figure}[t]
%     \centering
%     \begin{minipage}[c]{0.60\textwidth}
%         \includegraphics[width=\linewidth]{figures/teaser_v2.png}
%     \end{minipage}\hfill
%     \begin{minipage}[c]{0.37\textwidth}
%         \captionsetup{skip=0pt, justification=raggedright}%
%         \caption{Comparison of forecasting paradigms. (Top) TSFMs capture periodicity but miss exogenous shocks. (Middle) Standalone LLMs ingest text and numbers in one prompt, causing uncalibrated drift. (Bottom) \ours{} reconciles a numerical prior with macro/micro contextual forecasts via closed-form Bayesian synthesis. }
%         \label{fig:overview}
%     \end{minipage}
% \end{figure}

Time series forecasting is pivotal to decision-making across high-stakes domains such as finance, energy management, and supply chain logistics~\citep{lai2018modeling,zhou2021informer,mancuso2021machine,godahewa2021monash}. Recently, Time Series Foundation Models (TSFMs)~\citep{das2024decoder,goswami2024moment,woo2024moirai,ansari2024chronos,cohen2025time} have established a dominant paradigm. Pre-trained on massive corpora of numerical sequences, TSFMs capture complex temporal dependencies, periodicities, and multi-seasonal patterns. However, real-world systems are rarely closed. They are subject to exogenous qualitative shocks-geopolitical shifts, regulatory decisions, and supply interruptions-that precipitate structural breaks. Operating in a \textit{multimodal vacuum}, purely numerical TSFMs remain blind to these drivers because the governing signals lie outside historical numerical series.

\begin{wrapfigure}{r}{0.60\textwidth}
    \vspace{-9pt} % Pulls figure up to align cleanly with the top of the text
    \centering
    \includegraphics[width=\linewidth]{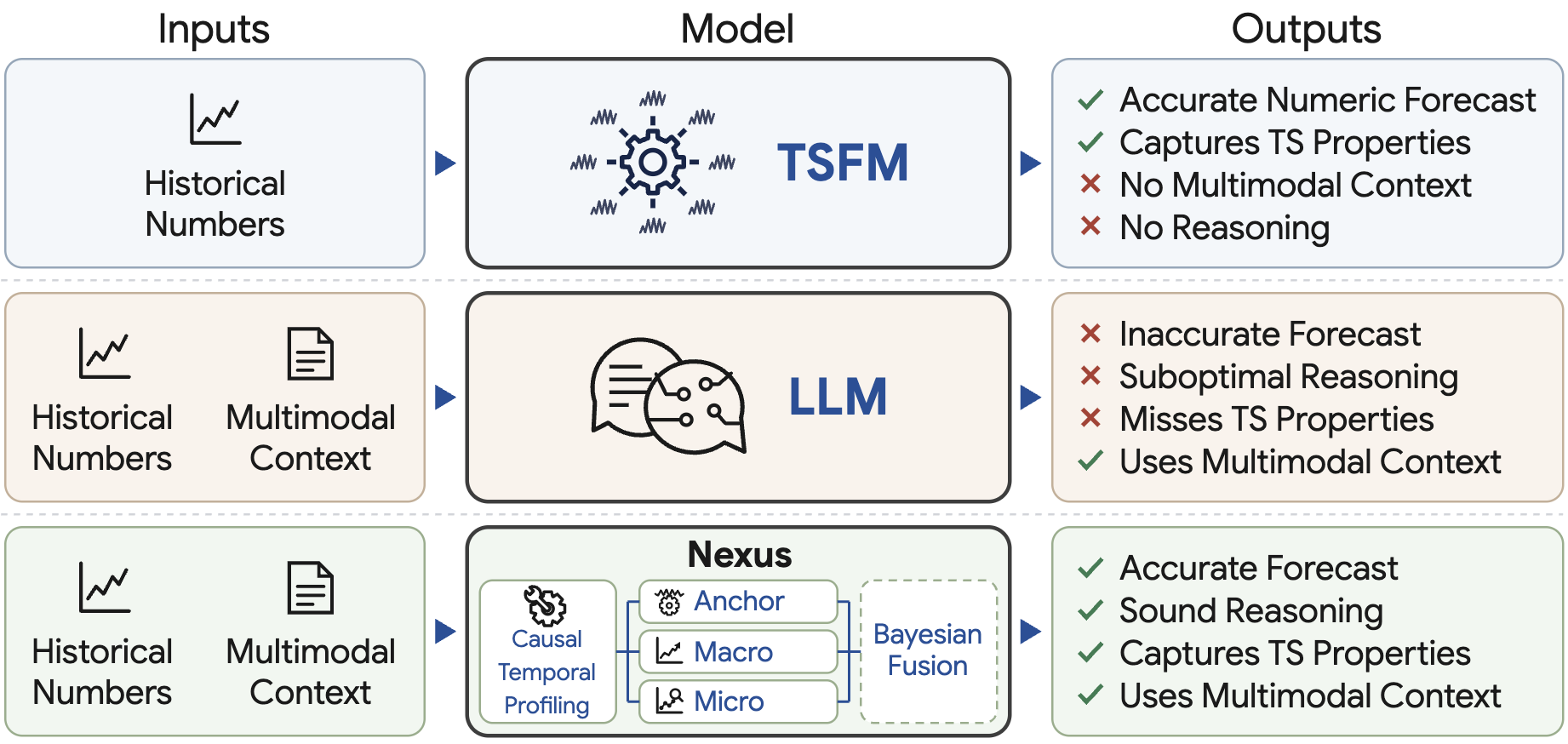}
    \captionsetup{skip=4pt, justification=raggedright}
    \caption{Comparison of forecasting paradigms. (Top) TSFMs capture periodicity but miss exogenous shocks. (Middle) Standalone LLMs ingest text and numbers in one prompt, causing uncalibrated drift. (Bottom) \ours{} reconciles a numerical prior with macro/micro contextual forecasts via closed-form Bayesian synthesis.}
    \label{fig:overview}
    \vspace{-9pt} % Tightens empty whitespace below the caption
\end{wrapfigure}

Conversely, Large Language Models (LLMs) excel at parsing unstructured textual narratives and evaluating causal real-world contexts~\citep{zhou2023onefitsall,kojima2022large,chen2025cctime,williams2025context}. Yet, standalone LLMs struggle with continuous numerical modeling and autoregressive sequence extrapolation~\citep{jin2024timellm,tan2024language}. Thus, current paradigms face a stark compromise: discard qualitative context to utilize statistical models, or rely on zero-shot LLM numerical generation prone to uncalibrated drift (Figure~\ref{fig:overview}). While recent efforts explore multimodal or tool-assisted workflows~\citep{das2025synapse,garza2025timecopilot,tao2026castr1}, they orchestrate external tools rather than resolving the statistical tension between semantic context and numerical extrapolation.

To bridge this, we introduce \textbf{\ours{}}, a multimodal forecasting framework that operates at the \textit{nexus} of structural numerical modeling and qualitative event reasoning. \ours{} works in three stages: i) \textit{temporal profiling and anchor generation}: an LLM analyzes past values and dated events to compile a causal-temporal profile, ii) \textit{dual contextual outlooks}: two streams query LLM across distinct temporal resolutions to generate decoupled, shape-first macro and catalyst-attuned micro trajectories calibrated on backtests; and iii) \textit{closed-form Bayesian synthesis}: we reconcile the numerical anchor with both contextual outlooks via Gaussian updating weighted by empirical error covariances. 

To evaluate real-world generalization while preventing data contamination, we benchmark \ours{} across eight datasets from three benchmarks: the living event-augmented \leaf{} benchmark~\citep{tan2026leaf} (hourly weather, electricity loads, and daily equity/commodity series evaluated after training cutoffs), the single-shock GIFT-CTX benchmark~\citep{salesforce_gift_ctx}, and the numerical GIFT-Eval benchmark~\citep{aksu2024gift}. With Gemini-3.1-Pro~\citep{gemini31pro} and Claude-Sonnet-4.5~\citep{claude_sonnet}, \ours{} consistently outperforms leading TSFMs (e.g., TimesFM-3.0, Chronos-2) and standalone LLM prompting baselines, reducing Weather MAE by $26\%$ and Electricity MAPE by $14\%$ over the strongest TSFM, while providing interpretable causal justifications preferred by independent evaluators (LLM-as-a-judge). Our primary contributions are as follows:
\begin{itemize}[leftmargin=*,itemsep=1pt,topsep=1pt]
    \item We demonstrate that LLMs achieve state-of-the-art predictive performance when contextual reasoning is decoupled from and systematically anchored to structural numerical baselines.
    \item We introduce \ours{}, a three-stage framework that couples diagnostic temporal profiling with multiscale (macro/micro) contextual outlooks and reconciles them with a numerical prior via closed-form Bayesian precision weighting.
    \item Across eight datasets spanning rich event contexts (\leaf{}), abrupt shocks (GIFT-CTX), and purely numerical series (GIFT-Eval), \ours{} outperforms foundation models and direct prompting baselines across LLM backbones.
\end{itemize}

\section{Problem Formulation}
\label{sec:problem_formulation}
We formulate the task of multimodal time series forecasting with explicit reasoning as jointly predicting the future values of a sequence and generating their underlying causal rationale, based on a multimodal observed historical context. Formally, let $\mathbf{X}_{1:\tau} = (x_1, x_2, \dots, x_\tau)$ represent a univariate numerical time series observed over a context window of length $\tau$. Concurrently, let $\mathbf{E}_{1:\tau} = (e_1, e_2, \dots, e_\tau)$ represent the associated unstructured textual data for each timestep in the context window. The complete historical context is thus the multimodal tuple $\mathcal{C}_{1:\tau} = (\mathbf{X}_{1:\tau}, \mathbf{E}_{1:\tau})$.

Given this context $\mathcal{C}_{1:\tau}$, the primary objective is to generate a numerical forecast for the subsequent $T$ timesteps, denoted as $\mathbf{X}_{\tau+1:\tau+T} = (x_{\tau+1}, x_{\tau+2}, \dots, x_{\tau+T})$. Crucially, unlike purely numerical forecasting, our goal is also to generate a natural language rationale $\mathbf{R}$ for the horizon, which explains the fundamental catalysts and events driving the predicted values.

Therefore, the problem can be formally framed as learning a mapping $\mathcal{F}$ that combines quantitative data and qualitative context into the predicted values and their justifications:
\[ \mathcal{F}(\mathbf{X}_{1:\tau}, \mathbf{E}_{1:\tau}) \rightarrow (\mathbf{X}_{\tau+1:\tau+T}, \mathbf{R}) \]

\section{Methodology: The \textsc{\ours} Framework}
\label{sec:methodology}

\begin{figure}[t]
    \centering
    \includesvg[width=\textwidth]{figures/nexus_arch_v6.svg}
    \caption{Overview of the \ours{} framework. First, an LLM conducts temporal profiling over numerical sequences and event records to establish a forecasting anchor (\textsection\ref{sec:phase1_prior}). Second, decoupled macro and micro streams yield backtest-calibrated contextual outlooks (\textsection\ref{sec:phase2_views}). Finally, the anchor prior and contextual views are combined in closed form via Bayesian covariance weighting (\textsection\ref{sec:phase3_bayesian}). Hexagons indicate LLM inference.}
    % \caption{Overview of the \ours{} framework. First, the backbone LLM analyzes the history and the event record to build a timestamped Causal-Temporal Profile $\mathcal{I}_{1:\tau}$, and selects a forecaster that produces the \textit{forecasting anchor} $\boldsymbol{\pi}$ with error covariance $\boldsymbol{\Sigma}$ (Section~\ref{sec:phase1_prior}). Second, two contextual streams query $f_{\text{LLM}}^{(\theta_m^*)}$ under stream-specific configurations $\theta_m^*$: the macro stream sketches the shape of the trajectory before assigning values, while the micro stream predicts the full horizon directly, both streams are calibrated on causal backtests (Section~\ref{sec:phase2_views}). $\mathcal{D}_{\text{tune}}$ refines each stream's configuration, $\mathcal{D}_{\text{val}}$ is held back to check it, and $\mathcal{D}_{\text{arb}}$, selects the final one. Finally, the three forecasts are combined in closed form, weighted by their backtest error covariances $\boldsymbol{\Sigma}$ and $\boldsymbol{\Omega}_m$ (Section~\ref{sec:phase3_bayesian}). Smooth hexagons denote LLMs.}
    \label{fig:nexus_architecture}
\end{figure}

% \subsection{Method Overview}
% \label{sec:method_overview}

While statistical models extrapolate historical regularities, reasoning over qualitative events is most valuable when such events affect the regular patterns. Rather than forcing one model to balance both modalities, \ours{} decouples structural numerical modeling from multiscale contextual reasoning, reconciling them via empirical error covariances in three stages (Figure~\ref{fig:nexus_architecture}). In the first stage of temporal profiling (\textsection\ref{sec:phase1_prior}), an LLM analyzes numerical sequences and dated event records to distill a \textit{causal-temporal profile}, committing a purely historical \textit{forecasting anchor} via a TSFM or a tailored forecasting model. In the dual contextual stage (\textsection\ref{sec:phase2_views}), two contextual streams query LLMs across distinct temporal resolutions, deriving a shape-first \textit{macro trajectory} and a catalyst, attuned \textit{micro outlook}, both calibrated on backtest errors. Finally, inspired by the Black-Litterman model~\citep{black1992global,idzorek2007step}, \ours{} reconciles these streams via closed-form Bayesian synthesis (\textsection\ref{sec:phase3_bayesian}). The numerical anchor serves as the prior, the contextual forecasts act as views, and the synthesized forecast is the posterior conditioned on empirical error covariances (Algorithm~\ref{alg:nexus}).

\subsection{Causal-Temporal Profiling and Forecasting Anchor Formulation}
\label{sec:phase1_prior}

\paragraph{Causal-Temporal Profiling.}
% Rather than feeding the entire multimodal history $\mathcal{C}_{1:\tau} = (\mathbf{X}_{1:\tau}, \mathbf{E}_{1:\tau})$ raw into a single prompt where long-range signals degrade \citep{liu2024lost}, 

We first conduct a causal temporal analysis by inspecting the series across multiple temporal scales via visual plots, characterizing sub-intervals numerically via programmatic change-point detection, querying the synchronous textual record through targeted search and date-range readers, and executing custom analysis scripts in a closed sandbox. By aligning historical regime changes with synchronous records in $\mathbf{E}_{1:\tau}$, the agent distills the history into a structured \textit{causal-temporal profile} $\mathcal{I}_{1:\tau}$ of three timestamped elements: (i) a \textit{segment} records its temporal span and the baseline behavior of the series within that interval,  (ii) a \textit{break} records regime switches, whether the textual record identifies a catalyst, and the trajectory response (step shift, momentary spike, or persistent drift), and (iii) a \textit{driver} records a recurring factor from the textual record alongside its direction, duration, and frequency.

% \begin{itemize}[leftmargin=*, topsep=2pt, itemsep=1pt]
%     \item A \textbf{segment} records its temporal span and the baseline regime behavior of the series within that interval.
%     \item A \textbf{break} records historical regime switches, whether the textual record identifies a catalyst, and the characteristic trajectory response (step shift, transient spike, or persistent drift).
%     \item A \textbf{driver} records a recurring trigger supported by the textual record alongside its direction, response latency, duration, and observed frequency.
% \end{itemize}
% Every element in $\mathcal{I}_{1:\tau}$ carries the timestamp by which it could have been known, enabling strict temporal clipping to $\mathcal{I}_{1:\tau_k}$ at any historical origin $\tau_k < \tau$.

\paragraph{Forecasting Anchor ($\boldsymbol{\pi}$).}
Guided by this analysis, the agent crafts a tailored prediction code in an isolated sandbox restricted to $\mathbf{X}_{1:\tau}$, or selects from a pool of pre-trained TSFMs (Chronos-2, TimesFM-2.5, and TiRex). It establishes a numerical prediction mapping $\mathcal{M}(\mathbf{X}_{1:\tau}) \rightarrow \boldsymbol{\pi}$ to produce the \textit{forecasting anchor} $\boldsymbol{\pi} = (\pi_{\tau+1}, \dots, \pi_{\tau+T})^\top \in \mathbb{R}^T$.  Appendices~\ref{appendix:profiling_prompt} and~\ref{app:tools} detail the instructions and tools used in this phase.

\begin{algorithm}[t]
\small % keeps every line of the algorithm on one line
\caption{\textsc{\ours}}
\label{alg:nexus}
\renewcommand{\algorithmicrequire}{\textbf{Input:}}
\renewcommand{\algorithmicensure}{\textbf{Output:}}
\begin{algorithmic}[1]
\Require History $\mathbf{X}_{1:\tau}$, dated events $\mathbf{E}_{1:\tau}$, horizon $T$, backbone language model $f_{\text{LLM}}$, scale $\tau_0$
\Ensure Synthesized forecast $\hat{\mathbf{X}}_{\tau+1:\tau+T}$ and rationale $\mathbf{R}$
\State \textbf{// Stage 1: Causal-Temporal Profiling and Forecasting Anchor (Section~\ref{sec:phase1_prior})}
\State Build the Causal-Temporal Profile $\mathcal{I}_{1:\tau}$ from $\mathcal{C}_{1:\tau} = (\mathbf{X}_{1:\tau}, \mathbf{E}_{1:\tau})$ 
\State Adopt a pretrained TSFM or craft a forecasting program $\mathcal{M}$
\State $\boldsymbol{\pi} \gets \mathcal{M}(\mathbf{X}_{1:\tau})$ \Comment{forecasting anchor}
% \State Anchor residuals at earlier origins $s$ with $s + T \le \tau$: $\mathbf{e}_s \gets \mathcal{M}(\mathbf{X}_{1:s}) - \mathbf{X}_{s+1:s+T}$
%\State $\boldsymbol{\Sigma} \gets (1-\lambda)\,\widehat{\boldsymbol{\Sigma}} + \lambda \operatorname{diag}(\widehat{\boldsymbol{\Sigma}}$, \quad where $\widehat{\boldsymbol{\Sigma}} = \operatorname{Cov}_s(\mathbf{e}_s)$
\State Replay $\mathcal{M}_{\text{prior}}$ across historical origins in $\mathbf{X}_{1:\tau}$ to estimate regularized anchor error covariance $\boldsymbol{\Sigma}$
\State \textbf{// Stage 2: Contextual Forecasts and Causal Backtest Calibration (Section~\ref{sec:phase2_views})}
\State Split backtest folds ending before $\tau$ into $\mathcal{D}_{\text{hist}} = \mathcal{D}_{\text{tune}} \cup \mathcal{D}_{\text{val}} \cup \mathcal{D}_{\text{arb}}$, with $\mathcal{D}_{\text{arb}}$ nearest to $\tau$
\For{each contextual stream $m \in \{\text{macro}, \text{micro}\}$}
    \For{each tuning fold $\tau_k \in \mathcal{D}_{\text{tune}}$, with the profile clipped to $\mathcal{I}_{1:\tau_k}$}
        \If{$m = \text{macro}$} \Comment{shape first}
            \State Sketch the shape $\mathcal{S}_{k}$, then assign values: $\hat{\mathbf{q}}_{m,k} \sim f_{\text{LLM}}^{(\theta_m)}(\mathcal{I}_{1:\tau_k}, \mathcal{C}_{\text{ctx},k}^{(\theta_m)}, \mathcal{S}_{k})$
        \Else \Comment{stepwise}
            \State Predict the full horizon stepwise directly: $\hat{\mathbf{q}}_{m,k} \sim f_{\text{LLM}}^{(\theta_m)}(\mathcal{I}_{1:\tau_k}, \mathcal{C}_{\text{ctx},k}^{(\theta_m)})$
        \EndIf
        \State Compare the residuals $\hat{\mathbf{q}}_{m,k} - \mathbf{X}_{\tau_k+1:\tau_k+T}$ with the events $\mathbf{E}_{\tau_k+1:\tau_k+T}$, refine $\theta_m$ by corrective feedback
    \EndFor
    \State Commit $\theta_m^*$ by arbitration on $\mathcal{D}_{\text{arb}}$ and check it on the held-back folds $\mathcal{D}_{\text{val}}$
    \State Per-lead error variances of $\theta_m^*$: $\sigma_{m,h}^2 \gets \operatorname{Var}_{k \in \mathcal{D}_{\text{hist}}}\!\big(\hat{q}_{m,k,h} - x_{\tau_k+h}\big)$, \quad $h = 1, \dots, T$
    \State $(\mathbf{q}_m, \mathbf{R}^m) \sim f_{\text{LLM}}^{(\theta_m^*)}(\mathcal{I}_{1:\tau}, \mathcal{C}_{\text{ctx}}^{(\theta_m^*)})$ at the forecast origin $\tau$
\EndFor
\State $\mathbf{C}_{\tau} \gets$ cross-lead correlation of both streams' backtest residuals with origin $\le \tau$ 
%($\mathbf{I}_T$ if too few)
\State $\boldsymbol{\Omega}_m \gets \operatorname{diag}(\boldsymbol{\sigma}_m)\,\mathbf{C}_{\tau}\,\operatorname{diag}(\boldsymbol{\sigma}_m)$ for both streams %\quad where $\mathbf{C}_{\tau} = (1-\lambda)\,\widehat{\mathbf{C}}_{\tau} + \lambda\,\mathbf{I}_T$
\State \textbf{// Stage 3: Bayesian Synthesis (Section~\ref{sec:phase3_bayesian})}
\State $\hat{\mathbf{X}}_{\tau+1:\tau+T} \gets$ Eq.~\ref{eq:precision_weighted} with $(\boldsymbol{\pi}, \tau_0\boldsymbol{\Sigma})$, $(\mathbf{q}_{\text{macro}}, \boldsymbol{\Omega}_{\text{macro}})$ and $(\mathbf{q}_{\text{micro}}, \boldsymbol{\Omega}_{\text{micro}})$ 
%\State Compose the rationale $\mathbf{R}$ from $\mathcal{C}_{1:\tau}$, $\mathbf{R}^{\text{macro}}$ and $\mathbf{R}^{\text{micro}}$
\State \Return $\hat{\mathbf{X}}_{\tau+1:\tau+T}$, $\mathbf{R} \equiv (\mathbf{R}^{\text{macro}}, \mathbf{R}^{\text{micro}})$
\end{algorithmic}
\end{algorithm}

\subsection{Contextual Outlook Generation}
\label{sec:phase2_views}

While the forecasting anchor $\boldsymbol{\pi}$ (Section~\ref{sec:phase1_prior}) extrapolates temporal regularities, it operates in a contextual vacuum and cannot anticipate regime shifts driven by qualitative catalysts in $\mathbf{E}_{1:\tau}$. To incorporate these signals, \ours{} queries the LLM $f_{\text{LLM}}^{(\theta_m^*)}$ through two contextual streams, macro and micro, each combining the causal-temporal profile $\mathcal{I}_{1:\tau}$ with calibrated context $\mathcal{C}_{\text{ctx}}$.

\paragraph{Macro-Level Outlook ($\mathbf{q}_{\text{macro}}$).}
The macro stream queries $f_{\text{LLM}}^{(\theta_{\text{macro}}^*)}$ under a top-down, \textit{shape-first} reasoning paradigm. Before generating numerical values, the model first commits to a structured trajectory specification $\mathcal{S}_{\text{macro}}$ defining expected total movement across the horizon $T$, turning-point locations, and whether each inflection reflects a recurring cycle or a quoted event catalyst. Conditioned on $\mathcal{S}_{\text{macro}}$, $f_{\text{LLM}}^{(\theta_{\text{macro}}^*)}$ instantiates the continuous trajectory $\mathbf{q}_{\text{macro}} \in \mathbb{R}^T$ and its causal narrative $\mathbf{R}^{\text{macro}}$ (detailed workflow and prompts in Appendix~\ref{appendix:macro_prompt}).

\paragraph{Micro-Level Outlook ($\mathbf{q}_{\text{micro}}$).}
The micro stream queries $f_{\text{LLM}}^{(\theta_{\text{micro}}^*)}$ in a \textit{direct stepwise} generation scheme attuned to immediate news shocks. Evaluating imminent drivers and localized deviations from historical equilibrium within its context span $\mathcal{C}_{\text{ctx}}^{(\theta_{\text{micro}}^*)}$, $f_{\text{LLM}}^{(\theta_{\text{micro}}^*)}$ predicts the entire horizon directly in a single stepwise pass to produce $\mathbf{q}_{\text{micro}} \in \mathbb{R}^T$, followed by step-specific justifications $\mathbf{R}^{\text{micro}}$ (Appendix~\ref{appendix:micro_prompt}).

\paragraph{Causal Backtest Calibration.}
To calibrate each stream's prompting configuration $\theta_m$ (the context $\mathcal{C}_{\text{ctx}}^{(\theta_m)}$ and failure-derived corrective instructions) without future leakage, $\mathcal{C}_{1:\tau}$ is partitioned once prior to execution into disjoint rolling backtest folds $\mathcal{D}_{\text{hist}} = \mathcal{D}_{\text{tune}} \cup \mathcal{D}_{\text{val}} \cup \mathcal{D}_{\text{arb}}$ shared across both streams (Appendix~\ref{app:folds}). On the tuning folds $\mathcal{D}_{\text{tune}}$, each calibration agent executes the forecaster using temporally clipped profiles $\mathcal{I}_{1:\tau_k}$ ($\tau_k < \tau$), inspects the resulting residuals against the held-out trajectory and synchronous events, and generates corrective feedback to refine $\theta_m$. Candidate configurations are then arbitrated on the origin-near folds $\mathcal{D}_{\text{arb}}$ preceding $\tau$ to commit $\theta_m^*$, evaluated for out-of-sample generalization on the held-back folds $\mathcal{D}_{\text{val}}$, and used across all evaluated folds in $\mathcal{D}_{\text{hist}}$ to estimate the stream's error covariance $\boldsymbol{\Omega}_m$.

\subsection{Bayesian Synthesis}
\label{sec:phase3_bayesian}

The final stage reconciles the anchor $\boldsymbol{\pi}$ with the two text-grounded forecasts ($\mathbf{q}_{\text{macro}}, \mathbf{q}_{\text{micro}}$) in closed form under Gaussian conjugacy \citep{bates1969combination, black1992global, he1999intuition}, weighting each source inversely to its empirical error covariance.

\paragraph{Empirical Covariance and Closed-Form Synthesis.}
We model the target trajectory $\mathbf{X}_{\tau+1:\tau+T} \in \mathbb{R}^T$ around the forecasting anchor $\boldsymbol{\pi}$ with error covariance $\boldsymbol{\Sigma} \in \mathbb{R}^{T \times T}$, and treat the contextual forecasts $\mathbf{q}_m$ as observations with error covariances $\boldsymbol{\Omega}_m \in \mathbb{R}^{T \times T}$. Both matrices are estimated from residuals observed before $\tau$. $\boldsymbol{\Sigma}$ comes from the anchor's own errors at earlier origins of the same series. $\boldsymbol{\Omega}_m$ holds the stream's per-lead error variances on $\mathcal{D}_{\text{hist}}$ together with a cross-lead correlation from the same series' earlier windows only, so that correlated errors are not taken independently (Appendix~\ref{app:covariance_details}). Under Gaussian conjugacy, the posterior trajectory $\hat{\mathbf{X}}_{\tau+1:\tau+T}$ has the closed-form solution:
\begin{equation}
    \hat{\mathbf{X}}_{\tau+1:\tau+T} = \left( (\tau_0 \boldsymbol{\Sigma})^{-1} +
    \boldsymbol{\Omega}_{\text{macro}}^{-1} + \boldsymbol{\Omega}_{\text{micro}}^{-1}
    \right)^{-1} \left( (\tau_0 \boldsymbol{\Sigma})^{-1} \boldsymbol{\pi} +
    \boldsymbol{\Omega}_{\text{macro}}^{-1} \mathbf{q}_{\text{macro}} +
    \boldsymbol{\Omega}_{\text{micro}}^{-1} \mathbf{q}_{\text{micro}} \right).
    \label{eq:precision_weighted}
\end{equation}

Intuitively, Equation~\ref{eq:precision_weighted} weights each forecast stream in proportion to its inverse error covariance (statistical confidence). Because $\boldsymbol{\Sigma}$ and $\boldsymbol{\Omega}_m$ preserve off-diagonal cross-lead correlations, the weights are not applied to each step $h$ in isolation, instead, the full trajectory is solved jointly, so reliable signal at one step informs neighboring leads. The scalar $\tau_0$ scales the prior covariance $\tau_0 \boldsymbol{\Sigma}$ and is fixed at $\tau_0 = 0.25$ (sensitivity analysis in Appendix~\ref{app:tau}).

% The weight each stream receives therefore varies along the horizon with its measured per-lead error; Appendix~\ref{app:covariance_details} reports the measured per-lead structure on the hourly cells. 

\providecommand{\todo}[1]{\textcolor{red}{\textbf{[TODO: #1]}}}
\providecommand{\best}[1]{\textbf{#1}}
\providecommand{\second}[1]{\underline{#1}}

\section{Experiments}
\label{sec:experiments}

We evaluate \ours{} across three distinct contextual regimes: dense multimodal event streams (\leaf{}), isolated qualitative shocks (GIFT-CTX), and context-free numerical series (GIFT-Eval). 
% In this section, we evaluate the forecasting accuracy and adaptability of \ours{} across three evaluation regimes spanning dense event streams, single-shock descriptions, and purely numerical time series. To begin, \textsection~\ref{sec:exp_setup} details the benchmarks, backbone models, baselines, and evaluation metrics. Subsequently, \textsection~\ref{sec:exp_leaf} presents our primary results on the multi-domain \leaf{} benchmark and analyzes backbone transferability between language models (Gemini-3.1-Pro and Claude-Sonnet-4.5). \textsection~\ref{sec:exp_giftctx} evaluates zero-shot shock adaptation on GIFT-CTX under short histories. Conversely, \textsection~\ref{sec:exp_gifteval} assesses purely numerical forecasting on GIFT-Eval in the absence of textual context. In \textsection\ref{sec:exp_reasoning}, we evaluate the reasoning of \ours{} with LLM-as-a-Judge with two separate LLM backbones (claude-sonnet-5 and claude-opus-5). Later, in Section~\ref{sec:components}, we isolate the constituent components of \ours{} to empirically analyze their performance characteristics. \todo{requires adding till Section 4.9 as we hope to have a lot of ablations}. 

\subsection{Experimental Setup}
\label{sec:exp_setup}

\paragraph{Benchmarks.}
To evaluate contextual forecasting across varying textual densities while mitigating data leakage, we evaluate on three benchmarks (Table~\ref{tab:datasets}, details in Appendix~\ref{app:datasets}): i) \textbf{LEAF} \citep{tan2026leaf} provides dense event records across domains like temperature and electricity, with data postdating September 2025 to rule out parametric leakage; ii) \textbf{GIFT-CTX} \citep{salesforce_gift_ctx} pairs numerical sequences with natural-language descriptions of isolated regime shifts and shock events; and iii) \textbf{GIFT-Eval} \citep{aksu2024gift} evaluates zero-shot numerical extrapolation without textual context across healthcare and sales domains.\footnote{GIFT-Eval sequences predate the pre-training cutoff of Gemini 3.1 Pro and Claude Sonnet 4.5, so numerical results should be interpreted with this parametric limitation in mind.}

\begin{table}[t]
\centering
\caption{Forecasting error on \leaf{} across shared evaluation windows. Weather reports MAE in $^\circ$C; electricity RMSE is in MW, market RMSE in price units. Gemini-3.1-Pro and Claude-Sonnet-4.5 is driven by the benchmark's own prompt with the event record, MoiraiAgent is the released selector on the input history. Lowest error in \best{bold}, second lowest \second{underlined}.}
\label{tab:leaf_main}
\small
\setlength{\tabcolsep}{4pt}
\resizebox{\textwidth}{!}{%
\begin{tabular}{l cc cc cc cc cc cc}
\toprule
& \multicolumn{2}{c}{\textbf{Weather} (72\,h)} & \multicolumn{2}{c}{\textbf{Electricity} (72\,h)} & \multicolumn{2}{c}{\textbf{Equities} (3\,d)} & \multicolumn{2}{c}{\textbf{Commodities} (3\,d)} & \multicolumn{2}{c}{\textbf{Equities} (7\,d)} & \multicolumn{2}{c}{\textbf{Commodities} (7\,d)} \\
\cmidrule(lr){2-3} \cmidrule(lr){4-5} \cmidrule(lr){6-7} \cmidrule(lr){8-9} \cmidrule(lr){10-11} \cmidrule(lr){12-13}
\textbf{Method} & MAE & RMSE & MAPE\,\% & RMSE & MAPE\,\% & RMSE & MAPE\,\% & RMSE & MAPE\,\% & RMSE & MAPE\,\% & RMSE \\
\midrule
\multicolumn{13}{l}{\textit{Supervised deep models}} \\
Autoformer                           & 3.506 & 4.677 & 10.125 & 2456.7 & 2.670 & 13.030 & 3.393 & 49.607 & 3.843 & 19.373 & 4.616 & 71.719 \\
TimesNet                             & 2.916 & 3.965 & 8.255 & 1910.1 & 1.991 & 10.468 & 2.585 & 33.897 & \best{2.948} & 15.414 & 3.653 & 50.687 \\
PatchTST                             & 3.003 & 4.064 & 10.156 & 2348.1 & 2.034 & 10.649 & 2.570 & 35.635 & 3.011 & 15.610 & 3.669 & 52.440 \\
iTransformer                         & 2.924 & 4.014 & 7.830 & 1859.8 & 2.031 & 10.603 & 2.570 & 34.398 & 2.971 & 15.351 & 3.619 & 50.520 \\
TimeMixer                            & 2.863 & 3.926 & 8.901 & 2149.9 & 2.020 & 10.586 & 2.538 & 34.300 & 2.993 & 15.416 & 3.907 & 54.241 \\
\midrule
\multicolumn{13}{l}{\textit{LLM-based supervised models}} \\
One-Fits-All (GPT4TS)                & 2.909 & 3.947 & 7.474 & 1780.5 & 1.989 & \second{10.400} & \second{2.531} & 34.205 & \second{2.956} & \second{15.342} & {3.580} & 50.005 \\
Time-LLM                             & 3.039 & 4.088 & 8.743 & 2151.1 & 2.205 & 11.344 & 2.947 & 43.726 & 3.044 & 15.682 & 3.890 & 59.028 \\
% Time-LLM + events                    & 3.027 & 4.109 & 8.263 & 2044.2 & -- & -- & -- & -- & -- & -- & -- & -- \\
\midrule
\multicolumn{13}{l}{\textit{Zero-shot foundation models}} \\
TimesFM-3.0                          & 2.687 & 3.794 & 6.471 & 1587.7 & 2.244 & 11.454 & 2.759 & 35.844 & 3.410 & 17.735 & 4.010 & 58.090 \\
TimesFM-2.5                          & 2.770 & 3.888 & 6.939 & 1689.9 & 2.245 & 11.432 & 2.755 & 37.407 & 3.235 & 16.332 & 3.757 & 56.004 \\
Chronos-2                            & 2.659 & 3.753 & 6.189 & 1524.6 & 2.046 & 10.638 & 2.628 & 34.428 & 3.115 & 16.045 & 3.762 & 54.467 \\
Chronos-Bolt                         & 2.683 & 3.799 & 6.342 & 1554.3 & 2.038 & 10.767 & 2.600 & 33.339 & 3.121 & 16.439 & 3.716 & 54.149 \\
TiRex                                & 2.675 & 3.800 & 6.697 & 1632.8 & 2.068 & 10.746 & 2.606 & 32.574 & 3.111 & 15.901 & 3.695 & 51.199 \\
Moirai-2.0                           & 2.705 & 3.763 & 6.518 & 1578.4 & 3.057 & 15.287 & 3.287 & 44.138 & 3.759 & 18.420 & 4.210 & 64.950 \\
\midrule
\multicolumn{13}{l}{\textit{Language models with the event record}} \\
Claude-Sonnet-4.5                    & 3.703 & 4.914 & 16.511 & 3852.4 & 2.489 & 12.187 & 2.618 & 35.763 & 3.325 & 16.647 & \second{3.537} & 51.540 \\
Gemini-3.1-Pro                       & 3.070 & 4.121 & 6.896 & 1666.7 & 2.145 & 10.651 & 2.565 & \best{30.691} & 3.410 & 16.766 & 3.930 & 50.983 \\
MoiraiAgent                          & {2.632} & {3.727} & \second{5.698} & {1412.9} & \second{1.976} & 10.554 & 2.543 & 31.806 & 3.030 & 15.472 & 3.654 & \second{48.628} \\
\textbf{\ours{}} (Claude-Sonnet-4.5) & \second{2.052} & \second{2.802} & {5.821} & \second{1390.6} & {2.070} & 10.678 & {2.543} & 34.097 & {3.069} & {16.132} & 3.668 & {50.731} \\
\textbf{\ours{}} (Gemini-3.1-Pro)   & \best{1.970} & \best{2.717} & \best{5.305} & \best{1309.9} & \best{1.958} & \best{10.272} & \best{2.488} & \second{31.461} & 2.994 & \best{15.341} & \best{3.434} & \best{48.610} \\
\bottomrule
\end{tabular}}
\end{table}
% -------------------------------------------------------------------------------------

\paragraph{Models.}
We employ Gemini-3.1-Pro~\citep{gemini31pro} as the primary LLM backbone for \ours{} and language-model baselines. To verify cross-family transferability, we additionally evaluate Claude-Sonnet-4.5~\citep{claude_sonnet} on \leaf{} across identical forecast horizons.  More recent models (e.g., Gemini-3.8-Flash, \mbox{Claude-Opus-5}) are not used in forecasts, as their training data cutoff overlaps \leaf{}'s evaluation period, effectively risking leakage of parametric knowledge during evaluation.

\paragraph{Baselines.}
We compare against four baseline families: (i)~\emph{Supervised deep models}: Autoformer \citep{wu2021autoformer}, TimesNet \citep{wu2023timesnet}, PatchTST \citep{nie2022time}, iTransformer \citep{liu2024itransformer}, and TimeMixer \citep{wang2024timemixer}, we \textbf{trained} these baselines per domain using the Time-Series-Library protocol \citep{wang2024deep} prior to the evaluation period, (ii)~\emph{LLM-based supervised models}: the parameter-efficient reprogramming architectures One-Fits-All \citep{zhou2023onefitsall} and Time-LLM \citep{jin2024timellm} - also supervised trained on the entire domain, (iii)~\emph{Zero-shot foundation models}: TimesFM-3.0 and TimesFM-2.5 \citep{das2024decoder}, Chronos-2 \citep{ansari2025chronos}, Chronos-Bolt \citep{ansari2024chronos}, TiRex \citep{auer2025tirex}, and Moirai-2.0 \citep{liu2025moirai}, and (iv)~\emph{Language models with event records}: Gemini-3.1-Pro and Claude-Sonnet-4.5 using \leaf{}'s official prompts, alongside MoiraiAgent. All baseline prompts are used from their benchmarks.
\setlength{\intextsep}{0pt}%
\begin{wraptable}[21]{r}{0.44\textwidth}
\centering
\captionsetup{justification=raggedright}%
\caption{Comparison results on GIFT-CTX, across all $245$ instances (lower is better). \ours{} runs on Gemini-3.1-Pro; the Gemini-3.1-Pro row is the benchmark's agent harness. $^\dagger$ denotes original GIFT-CTX Leaderboard entries. Lowest error in \textbf{bold}}
\label{tab:giftctx}
\small
\setlength{\tabcolsep}{3.5pt}
\begin{tabular*}{\linewidth}{l@{\extracolsep{\fill}}rr}
\toprule
\textbf{Method} & \textbf{wNMAE} & \textbf{wNRMSE} \\
\midrule
\multicolumn{3}{l}{\footnotesize\textit{Zero-shot foundation models}} \\
Chronos-2                  & 0.193 & 0.327 \\
MoiraiAgent$_\mathrm{sel}$ & 0.202 & 0.339 \\
TimesFM-3.0                & 0.208 & 0.353 \\
TiRex                      & 0.210 & 0.357 \\
TimesFM-2.5                & 0.218 & 0.363 \\
Moirai-2.0                 & 0.237 & 0.388 \\
Chronos-Bolt               & 0.268 & 0.441 \\
\midrule
\multicolumn{3}{l}{\footnotesize\textit{Language models with shock description}} \\
GPT-5.2$^\dagger$          & 0.247 & 0.376 \\
Claude-Opus-4.5$^\dagger$  & 0.171 & 0.262 \\
Gemini-3.0-Pro$^\dagger$   & 0.146 & 0.223 \\
MoiraiAgent$^\dagger$      & 0.124 & 0.184 \\
Gemini-3.1-Pro             & 0.068 & 0.103 \\
\midrule
% \textbf{\ours{}} micro     & \second{0.057} & \second{0.091} \\
% \textbf{\ours{}} macro     & \best{0.037} & \best{0.058} \\
\textbf{\ours{} } & \best{0.052} & \best{0.082}\\
\bottomrule
\end{tabular*}
\end{wraptable}
\paragraph{Metrics.}
On \leaf{}, we report MAPE and RMSE, substituting MAE ($^\circ$C) for MAPE on \textit{Weather} whenever temperatures cross zero. On GIFT-Eval, we report configuration-level MASE and the leaderboard aggregate (geometric mean of MASE relative to seasonal naive). For GIFT-CTX, we report the geometric mean of weighted normalized MAE and RMSE (wNMAE, wNRMSE), using original benchmark setting. Finally, in GIFT-Eval, we compare the MASE scores, as used in the original benchmark. Formal definitions are in Appendix~\ref{app:metrics}.
\setlength{\intextsep}{\nexusintextsep}% the class value again

\subsection{Event-Augmented Forecasting on \leaf{}}
\label{sec:exp_leaf}

As demonstrated in Table~\ref{tab:leaf_main}, \ours{} (Gemini-3.1-Pro) matches or outperforms the strongest baselines on every \leaf{} domain and horizon. The gains are most pronounced on the hourly physical domains, where dense event records interact with the periodic structure of the series. On Weather, \ours{} reduces MAE by $25\%$ relative to the strongest baseline (MoiraiAgent) and by $36\%$ relative to Gemini-3.1-Pro under the benchmark's own prompt. On Electricity, it lowers MAPE by $14\%$ relative to the best zero-shot foundation model (Chronos-2) and by $7\%$ relative to MoiraiAgent, which selects among the same models from their held-out errors.

In contrast, the daily Equities and Commodities domains are substantially more challenging, as asset prices closely approximate random walks over short horizons. Even in this low-signal regime, \ours{} attains the lowest MAPE on three-day equities and on both commodity horizons, outperforming supervised deep models that benefit from in-domain training across all market entities. Relative to Gemini-3.1-Pro under the benchmark's own prompt, it lowers MAPE by $3\%$ to $13\%$ on every market setting. %On seven-day equities, several supervised models obtain marginally lower MAPE, yet the gap to the best of them, TimesNet, is only $1.6\%$. \todo{one sentence on paired-window significance for weather and electricity once the bootstrap script has been run on the final window set}

\paragraph{Backbone Transferability.}
Table~\ref{tab:leaf_main} also shows that the gains of \ours{} carry over to Claude-Sonnet-4.5. Relative to the same backbone under the benchmark's prompt, \ours{} (Claude-Sonnet-4.5) reduces Weather MAE by $45\%$ and Electricity MAPE by $65\%$. Standalone prompting is strongly backbone-dependent on Electricity, where Sonnet's error is more than twice Gemini's ($16.5\%$ vs.\ $6.9\%$ MAPE), whereas inside \ours{} the two backbones land within 10\% of each other ($5.8\%$ vs.\ $5.3\%$).

\subsection{Single-Shock Adaptation on GIFT-CTX}
\label{sec:exp_giftctx}

% GIFT-CTX tests zero-shot adaptation when the context is a single shock description localized to a horizon region rather than a continuous event stream, and when the history is too short to form the disjoint backtest folds $\mathcal{D}_{\text{hist}}$. Without folds there is no error covariance for either contextual stream and no weight for the numerical prior, which cannot read the description in the first place. We therefore report the macro and micro streams of \ours{} separately rather than merging them by a weight the data cannot supply, nothing is tuned or selected on this benchmark.

% As shown in Table~\ref{tab:giftctx}, conditioning on the shock description allows language-model forecasters to reduce error by $3\times$ to $5\times$ relative to purely numerical foundation models, and both streams of \ours{} improve substantially on the benchmark's reference agent. 

%Interestingly, the two streams differ by a factor of $1.5$ on this benchmark even though no backtest exists to calibrate them; the weaker of the two still improves on every published entry. 

GIFT-CTX evaluates zero-shot adaptation when the context is a single shock description localized to a region of the forecast horizon. In addition, the numerical histories are too short to support causal temporal profiling or to construct the disjoint rolling backtest folds $\mathcal{D}_{\text{hist}}$ required for empirical calibration (Section~\ref{sec:phase2_views}). Consequently, the stream error covariances ($\boldsymbol{\Omega}_{\text{macro}}, \boldsymbol{\Omega}_{\text{micro}}$) cannot be estimated. \ours{} therefore falls back to an equal-weight average of the macro and micro outlooks. As shown in Table~\ref{tab:giftctx}, \ours{} achieves the lowest error among all evaluated methods, attaining a wNMAE of $0.052$ and a wNRMSE of $0.082$ across all instances. This represents a $73\%$ error reduction relative to the strongest zero-shot TSFM (Chronos-2, $0.193$ wNMAE) and a $58\%$ reduction over the published MoiraiAgent entry ($0.124$ wNMAE). This demonstrates that although equal weighting averages across the accuracy gap between the macro and micro trajectories, which cannot be identified without backtest folds, \ours{} provides a robust, parameter-free synthesis.

%Note that, these gains are obtained with a parameter-free equal weighting of the macro and micro forecasts, since no backtest folds exist to identify the accuracy gap between them. %(individual stream performance is reported in Appendix~\ref{app:giftctx_components}).

\subsection{Numerical Forecasting on GIFT-Eval}
\label{sec:exp_gifteval}

% Table 3 stands at the right of its own GIFT-Eval paragraph.
\setlength{\intextsep}{0pt}%
\begin{wraptable}[17]{r}{0.53\textwidth}
\centering
\captionsetup{justification=raggedright}%
\caption{GIFT-Eval short-term configurations, MASE (lower is better). \emph{Agg.} is the geometric mean of MASE relative to seasonal na\"ive across configurations; W, D and M denote weekly, daily and monthly. Lowest error in \best{bold}, second lowest \second{underlined}.}
\label{tab:gifteval}
\small
\setlength{\tabcolsep}{2.2pt}
\begin{tabular}{l cc c c c c}
\toprule
& \multicolumn{2}{c}{\textbf{Births}} & \textbf{COVID} & \begin{tabular}[b]{@{}c@{}}\textbf{Hier.}\\\textbf{Sales}\end{tabular} & \textbf{Hosp.} & \multirow{2}{*}{\textbf{Agg.}} \\
\cmidrule(lr){2-3} \cmidrule(lr){4-4} \cmidrule(lr){5-5} \cmidrule(lr){6-6}
\textbf{Method} & W & D & D & W & M & \\
\midrule
TimesFM-3.0      & 1.019 & \second{0.321} & \second{32.41} & \best{0.715} & \second{0.760} & \second{0.537} \\
Chronos-2        & 0.979 & 0.327 & 35.41 & \second{0.717} & 0.772 & 0.546 \\
Moirai-2.0       & \second{0.901} & 0.376 & 38.10 & 0.730 & 0.789 & 0.565 \\
TimesFM-2.5      & 1.087 & 0.343 & 36.91 & 0.720 & \best{0.757} & 0.566 \\
TiRex            & 1.071 & 0.409 & 38.94 & 0.722 & 0.762 & 0.592 \\
Chronos-Bolt     & 1.087 & 0.485 & 38.85 & 0.733 & 0.791 & 0.621 \\
\midrule
\textbf{\ours{}} & \best{0.854} & \best{0.316} & \best{29.84} & 0.748 & 0.761 & \best{0.513} \\
\bottomrule
\end{tabular}
\end{wraptable}
While \ours{} is designed for contextual forecasting, we evaluate it in a purely numerical setting ($\mathbf{E}_{1:\tau} = \emptyset$), with GIFT-Eval, on healthcare and sales domains (not covered in other datasets). While these datasets lack sufficient context for full stream calibration, it suffices for causal temporal profiling, so we adopt an equal-weight combination. Table~\ref{tab:gifteval} shows that even without textual context, \ours{} attains the lowest aggregate of every model measured ($0.513$ vs.\ $0.537$ for TimesFM-3.0) and the lowest error on three of the five configurations. Zero-shot TSFMs hold a marginal lead on hospital admissions and hierarchical sales, whose long, periodic histories favor numerical pattern extraction: here, our anchor matches the TSFMs, whereas the contextual streams add little non-periodic signal.
\setlength{\intextsep}{\nexusintextsep}% the class value again

% GIFT-Eval evaluates \ours{} in a purely numerical setting ($\mathbf{E}_{1:\tau} = \emptyset$), isolating the numerical prior of Section~\ref{sec:phase1_prior} and the contextual streams' reasoning over the series alone. 
% We report their equal-weight combination, Eq.~\eqref{eq:precision_weighted} with $\boldsymbol{\Sigma} = \boldsymbol{\Omega}_{\text{macro}} = \boldsymbol{\Omega}_{\text{micro}}$, which has no free parameter and applies one rule to every configuration, including the short COVID and hospital series that cannot support backtest folds.

% -------------------------------------------------------------------------------------

% -------------------------------------------------------------------------------------

\subsection{Reasoning Quality}
\label{sec:exp_reasoning}

Beyond numerical accuracy, we evaluate whether \ours{} also produces better reasoning than the standalone prompt. For each window, the macro and micro rationales are first pooled into a single rationale $R^{final} \gets f_{LLM}^{pool}(R^{macro}, R^{micro})$ (Appendix~\ref{app:reasoning_protocol}). Two independent judges, Claude-Sonnet-5 and Claude-Opus-5, then compare it blind against the rationale of Gemini-3.1-Pro under the benchmark's prompt. Each pair is judged in both presentation orders, and a side wins only if preferred in both, otherwise the pair is a tie. As shown in Table~\ref{tab:reasoning}, both judges prefer \ours{} overall ($71.08\%$ and $74.48\%$) and on every criterion, with the largest margin on analytical depth ($84.67\%$ and $80.48\%$), demonstrating its superiority in both accuracy and reasoning.

\begin{table}[t]
\centering
\caption{Blind pairwise reasoning evaluation on \leaf{}: wins / losses / ties of \ours{} against Gemini-3.1-Pro under the benchmark's prompt; equities and commodities pool both horizons, and the last column counts ties as half a win. Bold denotes winner. $^{\ast}$Neither arm retrieves external information, which keeps these margins more modest.}
\label{tab:reasoning}
\small
\setlength{\tabcolsep}{5pt}
\resizebox{\linewidth}{!}{%
\begin{tabular}{ll cccc c c}
\toprule
\textbf{Criterion} & \textbf{Judge} & \textbf{Weather} & \textbf{Electricity} & \textbf{Equities} & \textbf{Commodities} & \textbf{All} & \textbf{\ours{} score} \\
\midrule
\multirow{2}{*}{Domain relevance$^{\ast}$}   & Sonnet-5 & \textbf{286} / 2 / 332 & \textbf{108} / 10 / 335 & \textbf{527} / 26 / 2082 & \textbf{180} / 5 / 692 & \textbf{1101} / 43 / 3441 & \textbf{61.54\%} \\
                                    & Opus-5   & \textbf{433} / 7 / 180 & \textbf{209} / 37 / 207 & \textbf{805} / 227 / 1603 & \textbf{318} / 46 / 513 & \textbf{1765} / 317 / 2503 & \textbf{65.79\%} \\
\midrule
\multirow{2}{*}{Event relevance$^{\ast}$}    & Sonnet-5 & \textbf{252} / 50 / 318 & \textbf{92} / 48 / 313 & \textbf{1115} / 509 / 1011 & \textbf{363} / 228 / 286 & \textbf{1822} / 835 / 1928 & \textbf{60.76\%} \\
                                    & Opus-5   & \textbf{454} / 42 / 124 & \textbf{237} / 82 / 134 & \textbf{1521} / 712 / 402 & \textbf{474} / 256 / 147 & \textbf{2686} / 1092 / 807 & \textbf{67.38\%} \\
\midrule
\multirow{2}{*}{Internal coherence} & Sonnet-5 & \textbf{286} / 67 / 267 & \textbf{154} / 87 / 212 & \textbf{1245} / 149 / 1241 & \textbf{378} / 83 / 416 & \textbf{2063} / 386 / 2136 & \textbf{68.29\%} \\
                                    & Opus-5   & \textbf{258} / 8 / 354 & \textbf{206} / 6 / 241 & \textbf{849} / 86 / 1700 & \textbf{350} / 43 / 484 & \textbf{1663} / 143 / 2779 & \textbf{66.58\%} \\
\midrule
\multirow{2}{*}{Analytical depth}   & Sonnet-5 & \textbf{564} / 3 / 53 & \textbf{289} / 29 / 135 & \textbf{1945} / 146 / 544 & \textbf{627} / 68 / 182 & \textbf{3425} / 246 / 914 & \textbf{84.67\%} \\
                                    & Opus-5   & \textbf{588} / 5 / 27 & \textbf{304} / 56 / 93 & \textbf{1793} / 320 / 522 & \textbf{589} / 98 / 190 & \textbf{3274} / 479 / 832 & \textbf{80.48\%} \\
\midrule
\multirow{2}{*}{\textbf{Overall}}   & Sonnet-5 & \textbf{427} / 54 / 139 & \textbf{244} / 86 / 123 & \textbf{1638} / 494 / 503 & \textbf{472} / 214 / 191 & \textbf{2781} / 848 / 956 & \textbf{71.08\%} \\
                                    & Opus-5   & \textbf{572} / 17 / 31 & \textbf{321} / 66 / 66 & \textbf{1702} / 600 / 333 & \textbf{551} / 218 / 108 & \textbf{3146} / 901 / 538 & \textbf{74.48\%} \\
\bottomrule
\end{tabular}}
\end{table}

% -------------------------------------------------------------------------------------

% Table~\ref{tab:reasoning} shows that both judges prefer the rationale of \ours{} on a clear majority of windows, with the plain prompt preferred on $21\%$ and $14\%$ of windows respectively. The preference holds in every domain and is strongest on weather. The two judges agree on $64\%$ of windows and reach opposite verdicts on $8\%$. Per criterion, the margin is largest on analytical depth ($297$ vs.\ $16$ and $304$ vs.\ $34$ windows) and internal coherence, and smallest on event relevance, on which the judge sees only the horizon's events and cannot credit citations from the lookback record.

\subsection{Ablation of \ours{} Components}
\label{sec:components}

\begin{table}[t]
\centering
\caption{Contribution of each component and of calibration on the \leaf{} windows of Table~\ref{tab:leaf_main} (Gemini-3.1-Pro). The first three rows forecast with one component alone; the fourth combines all three with fixed equal weights and no calibration. Lowest error in \best{bold}, second lowest \second{underlined}.}
\label{tab:components}
\small
\setlength{\tabcolsep}{3pt}
\resizebox{\linewidth}{!}{%
\begin{tabular}{l cc cc cc cc cc cc}
\toprule
& \multicolumn{2}{c}{\textbf{Weather}} & \multicolumn{2}{c}{\textbf{Electricity}} & \multicolumn{2}{c}{\textbf{Equities} 3\,d} & \multicolumn{2}{c}{\textbf{Commod.} 3\,d} & \multicolumn{2}{c}{\textbf{Equities} 7\,d} & \multicolumn{2}{c}{\textbf{Commod.} 7\,d} \\
\cmidrule(lr){2-3} \cmidrule(lr){4-5} \cmidrule(lr){6-7} \cmidrule(lr){8-9} \cmidrule(lr){10-11} \cmidrule(lr){12-13}
\textbf{Component} & MAE & RMSE & MAPE & RMSE & MAPE & RMSE & MAPE & RMSE & MAPE & RMSE & MAPE & RMSE \\
\midrule
Forecasting anchor $\boldsymbol{\pi}$ & 2.678 & 3.784 & 6.100 & 1520.9 & \second{1.962} & 10.285 & \best{2.470} & \best{31.170} & \best{2.949} & \best{15.246} & 3.539 & 49.050 \\
Macro stream & 2.487 & 3.456 & 6.474 & 1570.6 & 2.171 & 11.196 & 2.777 & 35.755 & 3.301 & 16.925 & 3.737 & 57.922 \\
Micro stream & 2.585 & 3.595 & 6.288 & 1555.1 & 2.148 & 11.930 & 2.709 & 34.727 & 3.308 & 17.397 & 3.612 & \second{48.765} \\
\midrule
\ours{} w/o calibration & \second{2.057} & \second{2.843} & \second{5.333} & \best{1294.0} & 1.986 & \best{10.232} & 2.524 & 32.220 & 3.044 & 15.580 & \second{3.469} & 49.896 \\
\midrule
\textbf{\ours{}} & \best{1.970} & \best{2.717} & \best{5.305} & \second{1309.9} & \best{1.958} & \second{10.272} & \second{2.488} & \second{31.461} & \second{2.994} & \second{15.341} & \best{3.434} & \best{48.610} \\
\bottomrule
\end{tabular}}
\end{table}

To quantify the contribution of each component, Table~\ref{tab:components} evaluates the forecasting anchor. As demonstrated in the table, no single component consistently dominates, and their relative performance varies noticeably across domains. By synthesizing the macro and micro forecasts with the forecasting anchor, \ours{} attains the lowest overall error, reducing MAE by $21\%$ on Weather and MAPE by $13\%$ on Electricity relative to the best individual component. The market domains are considerably more difficult, as short-horizon prices closely follow a random walk and leave little exploitable signal for any component; even there, the synthesis remains within $1.5\%$ of the best component in every column, making it the most robust configuration across all domains.

\paragraph{Causal Backtest Calibration.} As discussed in Section \ref{sec:phase2_views}, causal backtest calibration is central to \ours{}. To quantify its importance, we evaluate \ours{} without it. Lacking calibration errors, \ours{} defaults to fixed equal weights for synthesis. Table~\ref{tab:components} demonstrates that removing calibration increases the error in every domain, raising Weather MAE by $4.4\%$ and the MAPE of the remaining domains by up to $1.7\%$. The benefit of calibration is largest on Weather, where backtest errors recur most consistently and make the instructions and learned weights most informative.

% \subsection{Error Correlation Analysis}
% \label{sec:error_corr}
% \todo{rewrite in more details}
% In this section, we examine the error geometry of the forecasters to understand how \ours{} improves the performance consistently, and how their errors correlate with one another. For every forecaster, first the forecast errors at each lead of each window is computed. We then correlate these errors for each pair of forecasters over all windows and leads of a domain. As demonstrated in Figure~\ref{fig:error_corr}, the TSFMs make nearly identical errors, correlating at $0.86$ to $0.94$ in every domain, as they all extrapolate the same numerical history. In contrast, macro and micro contextual streams of \ours{}, which read the event record, predict differently: on Weather and Electricity, their errors correlate with the TSFMs and with the forecasting anchor at only $0.31$ to $0.51$. On the other hand, for market domains, all forecasters are exposed to the same unpredictable price movements, so their errors remain strongly correlated. This explains why \ours{} achieves such large gains on Weather and Electricity in Table~\ref{tab:leaf_main}, and more modest gains on the market domains.

\subsection{\ours{}: Forecast Complementarity}
\label{sec:error_corr}

\begin{figure}[t]
\centering
\begin{minipage}[c]{0.58\textwidth}
\includegraphics[width=\linewidth]{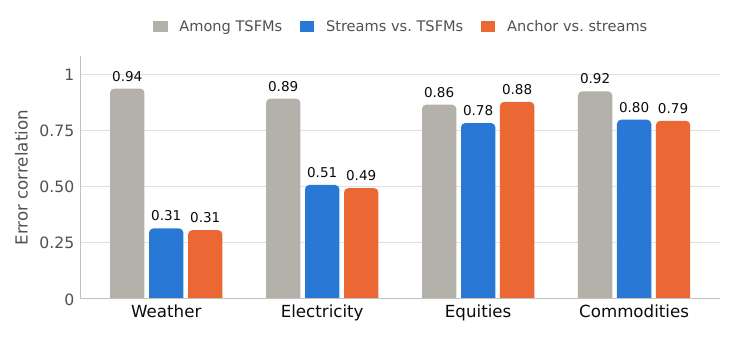}
\end{minipage}\hfill
\begin{minipage}[c]{0.39\textwidth}
\captionsetup{skip=0pt, justification=raggedright}%
\caption{Correlation between the errors of various forecasters on the windows of Table~\ref{tab:leaf_main}. Gray bars average over all pairs of zero-shot TSFMs, blue bars over each macro and micro stream of \ours{} paired with each TSFM, and orange bars over the forecasting anchor paired with each stream. Higher values mean that two forecasters tend to err in the same direction at the same time. Equities and Commodities pool both horizons.}
\label{fig:error_corr}
\end{minipage}
\end{figure}

To understand how \ours{} leverages multiple forecasting streams to improve accuracy (Table~\ref{tab:components}), we analyze the pairwise Pearson correlations of signed forecast errors across windows and leads in each domain (Figure~\ref{fig:error_corr}). Under Bayesian synthesis (Equation~\ref{eq:precision_weighted}), combining forecasts reduces error in proportion to how \emph{uncorrelated} their residuals are~\citep{bates1969combination}. Zero-shot TSFMs make nearly identical errors in every domain ($0.86$-$0.94$, gray bars), as they extrapolate the same numerical history. In contrast, \ours{}'s event-grounded macro and micro streams provide complementary signals: on Weather and Electricity, their errors correlate weakly with TSFMs ($0.31$-$0.51$, blue bars) and the forecasting anchor ($0.31$-$0.49$, orange bars), enabling synthesis to cancel errors and yield substantial gains. This also explains why gains on financial markets are more modest, as the accompanying text provides limited signal for unpredictable price movements, leaving errors more correlated across all forecasters ($0.78$-$0.88$).

\section{Related Works}

\paragraph{LLMs for Time Series Forecasting.} Recent work asks whether large language models (LLMs), trained on discrete text, can forecast continuous time series. Surveys identify directions such as direct prompting, numerical tokenization, modality alignment, and cross-modal bridging~\citep{zhang2024llmtsurvey}. \citet{gruver2023large} show that, with observations encoded as strings, LLMs can extrapolate zero-shot by next-token prediction in some settings. Other methods adapt pretrained transformers through lightweight modality alignment: GPT4TS/FPT~\citep{zhou2023onefitsall} transfers frozen language or vision transformers with limited parameter updates, TEMPO~\citep{cao2024tempo} adds inductive biases such as STL decomposition and prompt-based distribution adaptation, Time-LLM~\citep{jin2024timellm} reprograms patches into text prototypes and guides a frozen LLM with prompt-as-prefix conditioning, and UniTime~\citep{liu2024unitime} extends this direction to cross-domain multivariate forecasting. However, the utility of LLM backbones remains contested: \citet{tan2024language} show through ablations of LLM-based forecasters that removing or replacing the LLM often does not degrade performance, suggesting that much of the gain may come from patching, attention, or task-specific adaptation rather than linguistic pretraining. %\sarkar{add another sentence saying that new generation of reasoning models have not been investigated much in TS forecasting}

\paragraph{Time-Series Foundation Models.}
Another line pretrains forecasters only on temporal corpora, avoiding the text-modality gap. Chronos~\citep{ansari2024chronos} quantizes scaled values into a fixed vocabulary and trains transformer language models with cross-entropy loss for probabilistic zero-shot forecasting. TimesFM~\citep{das2024decoder} uses a patched decoder-only architecture for zero-shot forecasting across horizons and granularities, Lag-Llama~\citep{rasul2023lag} is a decoder-only probabilistic forecaster with lag covariates, MOMENT~\citep{goswami2024moment} learns general-purpose representations by masked reconstruction over the Time-series Pile, and MOIRAI~\citep{woo2024moirai} adds cross-frequency, any-variate, and mixture-distribution modeling for universal forecasting. These models show the promise of large-scale temporal pretraining, but generally remain static, single-pass predictors without explicit reasoning, revision, or use of external evidence.

\paragraph{Semantic, Adaptive, and Agentic Forecasting.}
Recent systems also use LLMs as semantic, adaptive, or agentic components. LoFT-LLM~\citep{you2025loftllm}, T-LLM~\citep{guo2026tllm}, and TimeSAF~\citep{zhang2026timesaf} study semantic calibration, temporal distillation, and asynchronous text-time-series fusion. Synapse~\citep{das2025synapse} arbitrates among TSFMs, TimeCopilot~\citep{garza2025timecopilot} coordinates feature analysis and model selection, and TimeSeriesScientist~\citep{zhao2025timeseriesscientist}, AlphaCast~\citep{zhang2025alphacast}, and Cast-R1~\citep{tao2026castr1} add multi-step planning, tool use, reflection, or memory. \citet{cheng2026agentic} argue for moving from static, model-centric prediction to iterative workflows of perception, planning, reflection, and memory. However, most of these systems operate over numerical histories, statistical diagnostics, model outputs, or tool-generated features. \ours{} instead grounds LLM reasoning in dated event text and weights each contextual forecast by its error on causal backtests.

\section{Conclusion}
% \label{sec:conclusion}
% In this paper, we introduced \ours, a novel multi-agent framework designed to tackle the complex challenge of multimodal contextual time-series forecasting. By decomposing forecasting into structured stages-Contextualization, Dual-Resolution Forecast Outlook Generation, and Forecast Synthesis and Calibration-\ours helps manage the complexity of processing long sequences of numerical data combined with unstructured text. \ours dynamically synthesizes broad macro-level trajectories with granular, event-driven micro-level catalysts, producing highly accurate numerical predictions alongside explicit reasoning. Through rigorous zero-shot evaluations on real-world Zillow real estate and Stock market datasets, we demonstrated that \ours consistently outperforms both time-series foundation model (TimesFM-2.5) and strong Chain-of-Thought LLM baselines. By bridging the gap between numerical trends and qualitative context, \ours offers a promising approach for developing interpretable, robust, and highly adaptable forecasting systems for complex, real-world domains.
%\todo{rewrite conclusion}
In this paper, we introduced \ours{}, a framework for multimodal contextual time series forecasting. Across three stages (Causal-Temporal Profiling, Contextual Outlook Generation, and Bayesian Synthesis), \ours{} updates a text-free forecasting anchor with two contextual forecasts, tracing the macro-level trajectory and the micro-level response to events, and weights each by its leakage-free causal backtest error. In zero-shot evaluations on \leaf{}, whose windows postdate the backbone's knowledge cutoff, \ours{} outperforms TSFMs, supervised deep models, and standalone LLM forecasters. It also attains the lowest error on GIFT-CTX and the lowest aggregate error on a GIFT-Eval subset, transfers from Gemini-3.1-Pro to Claude-Sonnet-4.5, and produces rationales that independent judges prefer on a clear majority of windows. By grounding LLM reasoning in a numerical anchor, \ours{} shows that LLMs can be strong forecasters when given genuine textual context.

% \subsection*{AI Use Statement}

% In this work, LLMs (primarily Gemini-3.1-Pro, with Claude-Sonnet-4.5 in ablations) serve as core backbone components of \ours{} to perform causal-temporal profiling, generate macro and micro contextual forecasts and rationales, and calibrate prompting configurations on historical backtest folds. We also used Claude-Sonnet-5 and Claude-Opus-5 within an LLM-as-a-Judge protocol to evaluate reasoning quality. Generative AI was not used to derive the theoretical framework, formulate the Bayesian synthesis equations, design the experimental methodology, or interpret the numerical results. Additionally, we used generative AI tools to edit the manuscript for grammar and readability. The authors revised all AI-assisted outputs and take full responsibility for the contents of this paper.

% \subsection*{Reproducibility Statement}
% \todo{fix the final conclusion}

% %Anonymized source code is provided as supplementary material. 
% The complete methodology is formalized in Algorithm~\ref{alg:nexus} and Section~\ref{sec:methodology}. Implementation details, covariance estimation, hyperparameters, and sensitivity analyses are detailed in Section~\ref{sec:experiments} and respective Appendices. Full agent tool specifications and prompt templates are documented in Appendices. Finally, benchmark descriptions (LEAF, GIFT-CTX, GIFT-Eval), preprocessing steps, evaluation metrics, and the LLM-as-a-Judge protocol are provided in Section~\ref{sec:experiments} and respective Appendices.

\section*{Limitations}

{\pretolerance=-1 % straight to the hyphenating pass, which avoids two loose lines
Running the full pipeline from scratch (cold start) requires executing one profiling agent, two forecasting agents, and a calibration pass over several backtest folds for each forecasting agent; therefore, \ours{} costs more than a zero-shot TSFM, which requires a single forward pass per forecast. In our experiments, each forecast window is calibrated independently; since the calibrated instructions are written to generalize across a series (Appendix~\ref{appendix:calibration_prompt}), they can be reused for later forecasts of that series, leaving only the agents and the closed-form synthesis to run.\par}

\section*{Ethical Considerations}
\ours{} combines numerical forecasting with reasoning over dated real-world events, and we evaluate it on weather, electricity load, equity and commodity prices, and healthcare and sales series. Event-aware forecasts can support decisions such as grid load planning and preparation for weather-driven demand, and the accompanying rationales let practitioners inspect why a forecast changed. At the same time, a fluent rationale can make an incorrect forecast appear credible, and on financial markets, where our gains are modest, acting on such forecasts could lead to economic losses. \ours{} is a research prototype: its forecasts and rationales are meant to inform human judgment rather than replace it, and they do not constitute financial or operational advice. All data we use are public benchmark series and public news events without personal information. 

\bibliography{custom}

\appendix
\newpage

\newpage

\section{Datasets}
\label{app:datasets}

Table~\ref{tab:datasets} summarizes every dataset used in Section~\ref{sec:experiments}. The three benchmarks were chosen because they sit at three points on one axis, how much language accompanies the numbers, and we describe each in turn.

{\setlength{\intextsep}{20pt plus 2pt minus 4pt}%
\begin{table}[ht]
\centering
\caption{Evaluation datasets. \emph{Windows} is the number of forecast windows scored in Table~\ref{tab:leaf_main} for \ours{} and all baselines. For \leaf{}, \emph{Text} counts the distinct timestamped event records inside the histories of the scored windows.}
\label{tab:datasets}
\small
\setlength{\tabcolsep}{4pt}
\resizebox{\textwidth}{!}{%
% Columns 3 and 8 are p{} (ragged-right) so long cells wrap instead of shrinking the table
\begin{tabular}{l l >{\raggedright\arraybackslash}p{6.0cm} r l r r >{\raggedright\arraybackslash}p{3.6cm}}
\toprule
\textbf{Benchmark} & \textbf{Domain} & \textbf{Entities} & \textbf{Series} & \textbf{Frequency} & \textbf{Horizon} & \textbf{Windows} & \textbf{Text} \\
\midrule
\multirow{8}{*}{\leaf{}}
 & Weather        & Air temperature, 10 US cities & 10 & hourly & 72\,h & 620 & 8{,}952 timestamped event records \\
 & Electricity    & Load, 3 regions (Boston, New York, San Francisco) & 3 & hourly & 72\,h & 453 & 3{,}986 timestamped event records \\
 & \multirow{2}{*}{Equities} & \multirow{2}{6.0cm}{30 US-listed large-cap equities} & 30 & daily (trading days) & 3\,d & 1{,}319 & \multirow{4}{3.6cm}{57{,}042 timestamped event records across the 40 market tickers} \\
 &                &                                                 & 30 & daily & 7\,d & 1{,}320 & \\
 & \multirow{2}{*}{Commodities} & \multirow{2}{6.0cm}{Brent, WTI, gold, silver, copper, natural gas, coffee, corn, soybeans, wheat} & 10 & daily & 3\,d & 440 & \\
 &                &                                                 & 10 & daily & 7\,d & 440 & \\
\midrule
\multirow{6}{*}{GIFT-CTX}
 & Anomaly        & 60 instances  & 60  & sec.\ to monthly & 15 to 240 & 60  & one description of the shock and its region \\
 & Phase change   & 60 instances  & 60  & sec.\ to monthly & 15 to 240 & 60  & one description of the regime change \\
 & Future event   & 125 instances & 125 & sec.\ to monthly & 15 to 240 & 125 & one description of an announced event \\
\midrule
\multirow{6}{*}{GIFT-Eval}
 & \multirow{3}{*}{US births} & \multirow{3}{6.0cm}{1 series} & 1 & monthly & 12 & 2  & none \\
 &                            &                               & 1 & weekly  & 8  & 14 & none \\
 &                            &                               & 1 & daily   & 30 & 20 & none \\
 & Hierarchical sales & 118 series & 118 & weekly  & 8  & 472 & none \\
 & Hospital           & 767 series & 767 & monthly & 12 & 767 & none \\
 & COVID deaths       & 266 series & 266 & daily   & 30 & 230 & none \\
\bottomrule
\end{tabular}}
\end{table}
}

\paragraph{\leaf{}.}
\leaf{} \citep{tan2026leaf} is a living benchmark that pairs numerical series with fact-checked, timestamped events, with forecast origins placed after the knowledge cutoff of the evaluated models. It covers hourly weather (ten cities) and electricity load (three regions), forecast $72$ steps ahead from origins between October 2025 and July 2026, and daily equity and commodity prices, forecast three and seven trading days ahead from origins between September 2025 and January 2026. We use the benchmark's own forecast windows.

\paragraph{GIFT-CTX.}
GIFT-CTX \citep{salesforce_gift_ctx} contains $245$ instances, $120$ real series from CIK and $125$ synthesized from GIFT-Eval, each paired with a single textual description of an anomaly, a phase change or an upcoming event. The description is localized to a region of the horizon that the metric weights more heavily.

\paragraph{GIFT-Eval.}
{\pretolerance=-1 % straight to the hyphenating pass, which avoids two loose lines
GIFT-Eval \citep{aksu2024gift} is the standard leaderboard for zero-shot numerical forecasting and provides no multimodal context, therefore it isolates the numerical capability of \ours{}. Since each forecast of \ours{} is a multi-step language-model run, evaluating the full leaderboard is computationally prohibitive, we therefore select short-term configurations at daily, weekly and monthly frequency, spanning sales, and healthcare domain. We exclude a configurations:  monthly US births, whose test split contains only two windows. On the remaining five configurations (Table~\ref{tab:datasets}) we use the leaderboard's windows, horizons and seasonal periods, and report MASE. \par}

\section{Metrics}
\label{app:metrics}

Throughout, $y_{\tau+h}$ denotes the realized value at lead $h$ of a window with origin $\tau$ and horizon $T$, and $\hat{y}_{\tau+h}$ the forecast. Every metric is computed per window and averaged over the windows of a cell.

\paragraph{\leaf{}.}
We report the benchmark's market metrics on every domain:
\begin{equation}
    \mathrm{MAPE} = \frac{100}{T}\sum_{h=1}^{T}\frac{|y_{\tau+h}-\hat{y}_{\tau+h}|}{|y_{\tau+h}|}, \qquad
    \mathrm{RMSE} = \sqrt{\frac{1}{T}\sum_{h=1}^{T}\left(y_{\tau+h}-\hat{y}_{\tau+h}\right)^2},
\end{equation}
where RMSE is reported in the units of the series (degrees Celsius, megawatts, or price). On weather, air temperature crosses zero and MAPE is undefined, so we report $\mathrm{MAE} = \frac{1}{T}\sum_h |y_{\tau+h}-\hat{y}_{\tau+h}|$ in its place.

\paragraph{GIFT-CTX.}

We adopt the official GIFT-CTX scoring protocol without modification. The evaluation is designed to emphasize specific regions of interest (e.g., anomalies or regime changes), denoted as $\mathcal{R} \subseteq \{1,\dots,T\}$. To achieve this, the benchmark applies a step-wise weighting scheme: for each horizon step $h$, the weight $w_h$ is set to $\alpha = 3$ if $h \in \mathcal{R}$, and $\beta = 1$ otherwise. These weights are then globally rescaled such that $\sum_{h=1}^{T} w_h = T$. 

Model performance is measured using the weighted normalized mean absolute error (wNMAE) and the weighted normalized root mean squared error (wNRMSE), defined as:
\begin{equation}
    \mathrm{wNMAE} = \frac{\frac{1}{T}\sum_{h=1}^{T} w_h\,|\hat{y}_{\tau+h}-y_{\tau+h}|}{\max_h y_{\tau+h}-\min_h y_{\tau+h}}, \qquad
    \mathrm{wNRMSE} = \frac{\sqrt{\frac{1}{T}\sum_{h=1}^{T} w_h^2\,(\hat{y}_{\tau+h}-y_{\tau+h})^2}}{\max_h y_{\tau+h}-\min_h y_{\tau+h}}
\end{equation}
The aggregate benchmark score is computed as the geometric mean of these metrics across all 245 instances.

% Crucially, the evaluation enforces strict fallback mechanisms for invalid model outputs: 
% \begin{itemize}[leftmargin=*, topsep=2pt, itemsep=1pt]
%     \item \textbf{Missing forecasts:} Imputed using the mean of the historical context.
%     \item \textbf{Truncated forecasts:} Padded forward using the model's last predicted value.
% \end{itemize}
% Because these fallbacks heavily penalize the final score, a model's generation coverage reported as the \emph{answered} column in Table~\ref{tab:giftctx} is a vital metric for contextualizing the performance of models that fail to achieve a 100\% response rate.

\paragraph{GIFT-Eval.}
For a configuration with seasonal period $m$, the mean absolute scaled error of a window is
\begin{equation}
    \mathrm{MASE} = \frac{\frac{1}{T}\sum_{h=1}^{T}|y_{\tau+h}-\hat{y}_{\tau+h}|}{\frac{1}{\tau-m}\sum_{t=m+1}^{\tau}|y_t-y_{t-m}|},
\end{equation}
following \citet{hyndman2006another}, averaged over the windows of the configuration. The leaderboard's aggregate divides each configuration's MASE by that of the seasonal na\"ive forecast and takes the geometric mean of the ratios across configurations; we report the same quantity over the configurations that are complete.

\providecommand{\todo}[1]{\textcolor{red}{\textbf{[TODO: #1]}}}

\section{Covariance Estimation and Prior Sensitivity}
\label{app:covariance_details}

\paragraph{Backtest folds $\mathcal{D}_{\text{hist}}$.}
\label{app:folds}

Prior to forecasting, the history preceding $\tau$ is partitioned into rolling backtest folds $\mathcal{D}_{\text{hist}} $, shared by macro and micro contextual streams. Consecutive target windows are spaced $T$ steps apart, and end strictly before $\tau$. The folds closest to $\tau$ form the arbitration set $\mathcal{D}_{\text{arb}}$, on which candidate configurations are compared under the most recent regime, while the earlier folds are divided equally into a tuning set $\mathcal{D}_{\text{tune}}$ and a held-back validation set $\mathcal{D}_{\text{val}}$.

Every residual behind $\boldsymbol{\Omega}_m$ comes from runs of the fixed configuration $\theta_m^*$: its runs on $\mathcal{D}_{\text{tune}}$ and $\mathcal{D}_{\text{arb}}$, and a fresh run on $\mathcal{D}_{\text{val}}$ after commitment, which, untouched by tuning or selection, gives an unbiased estimate of its error. Using all of the window's own folds keeps enough residuals per lead but can make $\boldsymbol{\Omega}_m$ slightly optimistic; since Equation~\ref{eq:precision_weighted} depends only on the relative scale of $\tau_0\boldsymbol{\Sigma}$ and $\boldsymbol{\Omega}_m$, the $\tau_0$ sweep in Table~\ref{tab:tau_sensitivity} also bounds a uniform under-estimate.

\paragraph{Error covariances $\boldsymbol{\Sigma}$ and $\boldsymbol{\Omega}_m$.}
Both covariances describe how forecast errors vary and co-vary across the $T$ leads, and both are
estimated only from errors observed before $\tau$: $\boldsymbol{\Sigma}$ from the forecasting
anchor's errors at earlier origins of the same series, and $\boldsymbol{\Omega}_m$ from the errors of
contextual stream $m$ on the backtest folds. Since a covariance over a horizon of $T$ steps has
$T(T{+}1)/2$ free parameters and usually only a limited set of past residuals can be computed, each
estimate $\mathbf{\beta} \in \{\boldsymbol{\Sigma}, \boldsymbol{\Omega}_m\}$ is shrunk toward its
diagonal \citep{ledoit2003honey},
\begin{equation}
    \mathbf{\beta} \;\leftarrow\; (1 - \lambda)\,\mathbf{\beta} \;+\; \lambda\,\operatorname{diag}(\mathbf{\beta}).
\end{equation}
We set $\lambda = 0.2$ throughout, as we observed little change in the
results until $\lambda > 0.5$. The estimation of $\boldsymbol{\Omega}_m$ differs slightly, because
every stream residual requires a full model run, so a single window provides only a few backtest
folds. These folds suffice for the per-lead variances
$\sigma_{m,h}^2 = \operatorname{Var}_k(\hat q_{m,k,h} - x_{\tau_k+h})$, which describe how the
committed configuration $\theta_m^*$ performs at each lead, but not for the correlation across
leads. Since ignoring this correlation would count the same error at neighboring leads multiple times and overweight the stream, we estimate it, denoted $\mathbf{R}_{i,\tau}$, from the backtest
residuals of both streams in all windows of series $i$ with origin at or before $\tau$; this
correlation mainly reflects the error dynamics of the series. The resulting
$\boldsymbol{\Omega}_m = \operatorname{diag}(\boldsymbol{\sigma}_m)\,\mathbf{R}_{i,\tau}\,\operatorname{diag}(\boldsymbol{\sigma}_m)$
involves no values observed after $\tau$. When too few past residuals are available to estimate this correlation, (e.g. the first window of hourly cells in \leaf{}), $\mathbf{R}_{i,\tau}$ is replaced by the identity.

\paragraph{Sensitivity to $\tau_0$}
\label{app:tau}

The scalar $\tau_0$ scales the error covariance of the forecasting anchor in the synthesis, and thereby sets how much the anchor is trusted relative to the two contextual forecasts. Throughout all our experiments, we fix $\tau_0 = 0.25$. Since $\boldsymbol{\Sigma}$ is estimated at origins before $\tau$ that may overlap with those used to select the anchor, it can be slightly optimistic; the sweep in Table~\ref{tab:tau_sensitivity} therefore rescales it over a 40-fold range ($\tau_0 \in [0.05, 2]$). Performance varies very little: pooled MASE changes by at most $3.4\%$, and the fixed value $\tau_0 = 0.25$ attains the lowest pooled error. Finally, when the profile is replayed at a backtest origin $\tau_k$, every element dated after $\tau_k$ is removed and the free-text summaries written with the full history are withheld. Note that these guards concern calibration only, the evaluated forecast uses no information after $\tau$.

\begin{table}[t]
\centering
\caption{Sensitivity of \ours{} to $\tau_0$, measured by pooled MASE over all samples in Table~\ref{tab:leaf_main} (lower is better). The reported setting, $\tau_0 = 0.25$, was fixed before these numbers were produced. Lowest value is shown in \textbf{bold}.}
\label{tab:tau_sensitivity}
\small
\begin{tabular}{lcccccc}
\toprule
$\tau_0$ & 0.05 & 0.10 & 0.25 & 0.50 & 1.00 & 2.00 \\
% \midrule
% Weather           & 0.984 & 0.929 & 0.875 & \textbf{0.859} & 0.863 & 0.882 \\
% Electricity       & 0.876 & 0.853 & 0.832 & \textbf{0.831} & 0.842 & 0.862 \\
% Equities 3\,d     & 1.567 & \textbf{1.565} & 1.572 & 1.587 & 1.606 & 1.624 \\
% Commodities 3\,d  & \textbf{1.597} & 1.604 & 1.624 & 1.646 & 1.671 & 1.695 \\
% Equities 7\,d     & \textbf{2.356} & 2.361 & 2.385 & 2.417 & 2.454 & 2.487 \\
% Commodities 7\,d  & 2.661 & \textbf{2.657} & 2.660 & 2.678 & 2.705 & 2.732 \\
\midrule
Pooled MASE          & 1.674 & 1.661 & \textbf{1.658} & 1.670 & 1.690 & 1.714 \\
\bottomrule
\end{tabular}
\end{table}

% The individual domains reveal a consistent trade-off. On weather and electricity, where the contextual forecasts improve most on the anchor, a slightly larger value ($\tau_0 = 0.50$) performs best, whereas on equities and commodities, where price changes over a few days are difficult to anticipate, smaller values that lean on the anchor perform best. The fixed value lies between these two regimes and remains within $1.9\%$ of the best setting in every domain. Even at $\tau_0 = 2.00$, \ours{} attains a pooled MASE of $1.714$ and remains ahead of the strongest baseline in Table~\ref{tab:leaf_mase}.

\section{Agent Tool Surface}
\label{app:tools}

The agents act through a sandboxed tool interface rather than emitting numbers directly. We
describe that interface as the eight operations in Table~\ref{tab:tools}; each is exposed as a
small number of variants differing only in arguments and return shape.

\begin{table}[t]
\centering
\caption{Tools used by the agents.}
\label{tab:tools}
\small % 9pt, like the other tables
\begin{tabular}{l>{\raggedright\arraybackslash}p{9.4cm}}
\toprule
\textbf{Tool} & \textbf{Effect} \\
\midrule
\texttt{inspect\_series} & Summarize a span, locate change points, render the series as a
figure, and search the attached text records. \\
\texttt{run\_code} & Execute analysis code against the observed history in an isolated
process. \\
\texttt{build\_prior} & Enumerate candidate numerical predictors, score them at stored origins,
and commit one. The prior is frozen once committed and is never refitted. \\
\texttt{make\_backtests} & Materialize the causal folds $\mathcal{D}_{\text{hist}}$. \\
\texttt{write\_profile} & Record findings and finalize the Causal-Temporal Profile
$\mathcal{I}$. \\
\texttt{run\_forecaster} & Run a contextual forecaster under a candidate configuration
$\theta_m$ and return its trajectory and reasoning. \\
\texttt{evaluate\_attempt} & Score a returned trajectory on the folds, compare configurations,
inspect the reasoning behind an attempt, and set the standing instruction carried in
$\theta_m$. \\
\texttt{commit\_config} & Commit the final configuration $\theta_m^*$. \\
\bottomrule
\end{tabular}
\end{table}

All tools read from $\mathcal{C}_{1:\tau}$ only. No tool exposes values at or beyond the
forecast origin, and at a historical fold origin $\tau_k$ every tool is restricted to
$\mathcal{C}_{1:\tau_k}$.

\section{Agent Prompts}
\label{app:prompts}

We provide the system prompts and user templates used for each agent in the \ours{} framework.
Larger instructions are abridged: each retains the contract that determines the agent's output, and omits
formatting boilerplate and error-recovery text.

\subsection{Causal-Temporal Profiling Agent}
\label{appendix:profiling_prompt}
\begin{tcolorbox}[title=Causal-Temporal Profiling Agent Prompts, breakable]
\small\ttfamily
\textbf{System Prompt:}\\*
You are an expert causal analysis agent. You analyse one time series together with the text
records attached to it, using the observed history only.

\vspace{4pt}
\textbf{User Template:}\\*
\textbf{Task:}\\*
Establish, from \{history\_str\} alone, the regimes "\{target\_name\}" has passed through and
what separated them; which recurring patterns hold and at what period; and which text records
moved the series and which did not. Plot the series and read the figure before you conclude.

\textbf{Domain:} \{domain\} \quad \textbf{Horizon:} \{horizon\}

\textbf{Historical Data (Text \& Values):}\\*
\{history\_str\}

\textbf{Then commit the numerical prior.}\\*
Enumerate the candidate predictors available to you, score them at the provided backtest
origins, and plot the leading ones against the observed continuation before you choose. Select
on that evidence. Where no candidate suits the series, write a closed-form
predictor instead and state the assumption it encodes.

\textbf{Output:}\\*
A structured profile in which every element carries the timestamp by which it could have been
known, followed by the committed predictor.
\end{tcolorbox}

\subsection{Macro-Reasoning Forecaster}
\label{appendix:macro_prompt}
The contextual forecaster invoked by the macro calibration agent under a candidate configuration
$\theta_{\text{macro}}$. It runs as turns of one stateful thread: the shape is fixed before
any number exists, the numbers are filled into that shape, and the rationale is recovered last.
The final turn inherits the model's reasoning state from turn one rather than re-deriving it,
so the thread is carried forward rather than re-sent.

\begin{tcolorbox}[title=Macro-Reasoning Forecaster Prompts, breakable]
\small\ttfamily
\textbf{System Prompt:}\\*
You are a contextual numerical forecasting agent. You reason about trajectory shape before
committing to numbers.

\vspace{6pt}
\textbf{Turn 1 (shape).}\\*
\textbf{Target:} \{target\_name\} \quad \textbf{Domain:} \{domain\}\\
\textbf{Forecast origin:} \{origin\} \quad \textbf{Horizon:} \{horizon\} steps, one step =
\{frequency\}\\
\textbf{Profile:} \{profile\} \quad \textbf{Numerical context:} \{context\_str\}

Before predicting any numbers, decide the SHAPE of the next \{horizon\} steps. Answer briefly
and concretely:\\
1. Direction: over the whole horizon, is the series going up, down or sideways, and by roughly
how much in total, in the units of the series?\\
2. Turning Points: is the horizon one continuous movement, or does it change somewhere? If it
changes, say at which step and what changes. If nothing changes, say so; a flat verdict is a
real verdict.\\
3. Why each turning point, pattern, or event? Answer per turning point, not in general. A claim
of an event catalyst must quote a record that predates the forecast origin.\\
4. Falsification: what would have to happen for this shape to be wrong?

\vspace{6pt}
\textbf{Turn 2 (fill), stateful same thread.}\\*
Now fill in that shape with numbers, for these timestamps: \{future\_dates\}.\\
Follow the shape you just described: the direction, the turning points and the mechanism. If
writing the numbers convinces you the shape was wrong, put what changed inside
\texttt{<revision>} tags BEFORE the prediction, then give the numbers you actually believe.\\
Output exactly \{horizon\} numbers, comma separated, inside \texttt{<prediction>} tags, and put
nothing else inside those tags.

\vspace{6pt}
\textbf{Output Format:}\\*
Turn 1: the four answers above.\\
Turn 2: \texttt{<revision>} optional \texttt{</revision>} then
\texttt{<prediction>} exactly \{horizon\} values \texttt{</prediction>}\\
\texttt{<reasoning>} $\mathbf{R}^{\text{macro}}$, the causal narrative \texttt{</reasoning>}
\end{tcolorbox}

\needspace{10.8\baselineskip}  % keep the heading with at least the System Prompt of its prompt box
\subsection{Micro-Reasoning Forecaster}
\label{appendix:micro_prompt}
The contextual forecaster invoked by the micro calibration agent under a candidate configuration $\theta_{\text{micro}}$.

\begin{tcolorbox}[title=Micro-Reasoning Forecaster Prompts, breakable, lines before break=2]
\small\ttfamily
\textbf{System Prompt:}\\*
You are a contextual numerical forecasting agent attuned to imminent catalysts and to short-run
deviations from the historical norm.

\vspace{4pt}
\textbf{User Template:}\\*
\textbf{Task:}\\*
Predict the next \{horizon\} values for "\{target\_name\}" in a single pass.

\textbf{Causal-Temporal Profile:} \{profile\}\\
\textbf{Numerical Context:} \{context\_str\}

\textbf{Instructions:}\\*
Attend to imminent catalysts and to deviations of the recent window from the historical norm.
Do not reason about long-range regime structure; it is already summarised in the profile.
Predict the full horizon directly, then justify the individual steps that depart from the
recent level.

\textbf{Output Format:}\\*
\texttt{<forecasted\_values>} exactly \{horizon\} values \texttt{</forecasted\_values>}\\
\texttt{<reasoning>} step-level justification \texttt{</reasoning>}
\end{tcolorbox}

\subsection{Calibration Prompts}
\label{appendix:calibration_prompt}
One per contextual stream. It does not forecast; it invokes the forecaster, evaluates what comes
back on the causal folds, and commits a configuration.

\begin{tcolorbox}[title=Calibration Prompts, breakable]
\small\ttfamily
\textbf{System Prompt:}\\*
You are a forecasting strategy optimiser. You do not produce forecasts yourself. You invoke a
forecaster under different configurations, evaluate it against outcomes that have been revealed
for historical folds only, and commit the configuration you will stand behind at the forecast
origin.

\vspace{4pt}
\textbf{User Template:}\\*
\textbf{Task:}\\*
Commit a configuration for the \{stream\} forecaster on "\{target\_name\}".

\textbf{Causal-Temporal Profile:} \{profile\}\\
\textbf{Tuning folds:} \{tune\_folds\} \quad \textbf{Held-back folds:} \{val\_folds\}\\
\textbf{Arbitration folds:} \{arb\_folds\}

\textbf{Procedure:}\\*
1. Run the forecaster on the tuning folds under a candidate configuration: the span of numerical
history to expose, whether to interleave the text records, whether to include the profile, and
any standing instruction.\\
2. Score what it returns, and plot it against the revealed outcome. Read the figure and the
agent's reasoning before you judge the numbers.\\
3. Reason where it failed and why, write a standing instruction that would have helped before the outcome was
visible. Ensure generalizability of the instructions across the time series.\\
4. Repeat for further candidates, then plot them together and compare them.\\
5. Commit on the arbitration folds nearest the origin, and say which windows your choice won and
by how much. The committed configuration is then scored once on the held-back folds.

\textbf{Output Format:}\\*
\texttt{<committed\_config>} the prompt configuration and the evidence for it
\texttt{</committed\_config>}
\end{tcolorbox}

\section{Reasoning Evaluation Protocol}
\label{app:reasoning_protocol}

\paragraph{Rationale of \ours{}.}
The macro and micro streams each return a rationale with their forecast. For the evaluation in Section~\ref{sec:exp_reasoning}, we pool the two into one final rationale per window. In the prompt structure below, the macro and micro rationales appear as Analyst~1 and Analyst~2.

\begin{tcolorbox}[title=Rationale Pooling Prompt, breakable]
\small\ttfamily
\textbf{System Prompt:}\\*
You are the forecaster that covers \{domain\_desc\}. Two other analysts have each written a rationale for the next \{horizon\} steps, from the same record. You have the full history: every observed value with its timestamp and the complete dated record of events. Write the desk's final reasoning: the most likely course of the series over the horizon and the drivers behind it.

\vspace{6pt}
You are given:\\
1. THE FULL HISTORY: \{history\_str\}\\
2. ANALYST 1 (outlook): committed to the overall shape of the horizon before writing numbers; their rationale.\\
3. ANALYST 2 (step): reasoned step by step; their rationale.

\vspace{6pt}
Weigh the two analysts against the full record, and find the reasoning trajectory supported by the history.

\vspace{6pt}
Structure, with bold labels: \textbf{Outlook} (two or three sentences), \textbf{What the record says} (dated bullets, one per event or source, in date order), \textbf{How the horizon unfolds} (in order, by step or group of steps, each with the driver behind it), \textbf{Where the analysts differed and which reading the record supports}, \textbf{Key risk}.

\vspace{6pt}
\textbf{User Prompt:}\\*
\textbf{Target:} \{target\_name\} for \{entity\}\\
\textbf{Forecast origin:} \{origin\}\\ \textbf{Horizon:} \{horizon\} steps, \{first\_date\} to \{last\_date\}\\
\textbf{The full history.} Values (timestamp, value): \{full\_history\_values\}\\
Dated record of events (date, source, description): \{full\_history\_record\}\\
\textbf{Analyst 1 (outlook):} \{macro\_rationale\}\\
\textbf{Analyst 2 (step):} \{micro\_rationale\}\\
Write the reasoning now.

\vspace{6pt}
\textbf{Output Format:}\\*
\textless{}reasoning\textgreater{} final rationale \textless{}/reasoning\textgreater{}
\end{tcolorbox}

\paragraph{Comparison arm.}
The comparison rationale is the reasoning text returned by Gemini-3.1-Pro under \leaf{}'s official prompt on the same window.

\needspace{10\baselineskip}  % keep the lead-in with its prompt box
\paragraph{Judge.}
The judge prompt is as below:

\begin{tcolorbox}[title=LLM-as-a-Judge Prompt, breakable]
\small\ttfamily
\textbf{System Prompt:}\\*[4pt]
You are an expert financial and time-series forecasting judge. Your task is to compare the reasoning quality of two different forecasting models and determine which one is better.\\[6pt]
You will be provided with:
\begin{enumerate}
    \setlength{\itemsep}{0pt}
    \setlength{\parskip}{0pt}
    \item GROUND TRUTH EVENTS: The actual events that occurred during the forecast horizon.
    \item MODEL A REASONING: The reasoning text from Model A.
    \item MODEL B REASONING: The reasoning text from Model B.
\end{enumerate}

You must evaluate both models based on the following criteria:
\begin{enumerate}
    \setlength{\itemsep}{0pt}
    \setlength{\parskip}{0pt}
    \item Domain Relevance: Correct use of domain-specific terminology (e.g., 'support levels', 'seasonality', 'macroeconomic headwinds').
    \item Event Relevance \& Plausibility: Does the reasoning logically and causally link the provided Ground Truth Events to the fluctuations it describes? Is the narrative highly plausible given the events? (Penalize hallucinations or illogical connections).
    \item Internal Coherence: Is the overall reasoning coherent?
    \item Analytical Depth: Does the reasoning demonstrate a deep understanding of fundamental dynamics rather than just superficial observations?
\end{enumerate}

Compare Model A and Model B. Decide which model provides better reasoning overall based purely on logical coherence, depth, and plausibility. If they are equally good or equally bad, you can declare a Tie.\\[6pt]
You must output your evaluation strictly as a JSON object matching the following schema exactly:\\
\{\\
\hspace*{2em}"domain\_relevance\_winner": "<Model A, Model B, or Tie>",\\
\hspace*{2em}"event\_relevance\_winner": "<Model A, Model B, or Tie>",\\
\hspace*{2em}"internal\_coherence\_winner": "<Model A, Model B, or Tie>",\\
\hspace*{2em}"analytical\_depth\_winner": "<Model A, Model B, or Tie>",\\
\hspace*{2em}"overall\_preference": "<Model A, Model B, or Tie>",\\
\hspace*{2em}\parbox[t]{\dimexpr\linewidth-2em\relax}{"justification": "<2-4 sentence explanation of why the preferred model is better overall, referencing specific criteria>"\strut}\\
\}

\vspace{8pt}
\textbf{User Template:}\\*[4pt]
--- GROUND TRUTH EVENTS ---\\
\{ground\_truth\_events\}\\[6pt]
--- MODEL A REASONING ---\\
\{model\_a\_reasoning\}\\[6pt]
--- MODEL B REASONING ---\\
\{model\_b\_reasoning\}\\[6pt]
Evaluate the reasoning of both models and output the strict JSON object with your preference.
\end{tcolorbox}

\section{Qualitative Forecast Examples}
\begin{figure}[H]
\centering
\begin{subfigure}{\linewidth}
\centering
\includegraphics[width=0.9\linewidth]{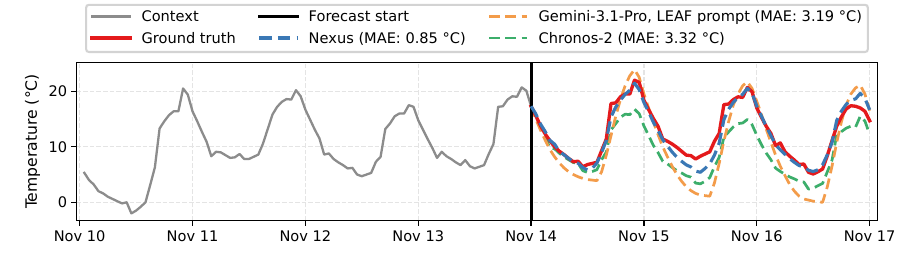}
\caption{Weather, Denver, 72\,h from 2025-11-14}
\end{subfigure}\\[8pt]
\begin{subfigure}{\linewidth}
\centering
\includegraphics[width=0.9\linewidth]{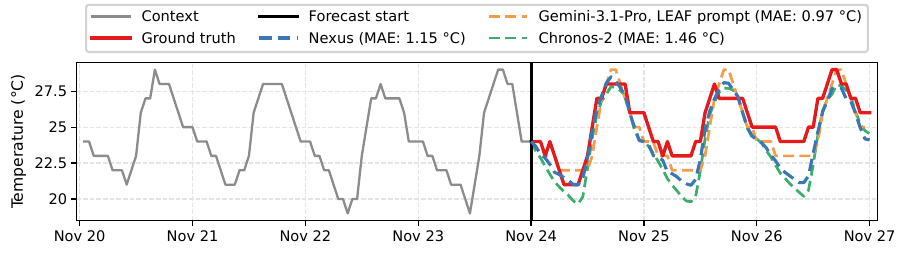}
\caption{Weather, Miami, 72\,h from 2025-11-24}
\end{subfigure}\\[8pt]
\begin{subfigure}{\linewidth}
\centering
\includegraphics[width=0.9\linewidth]{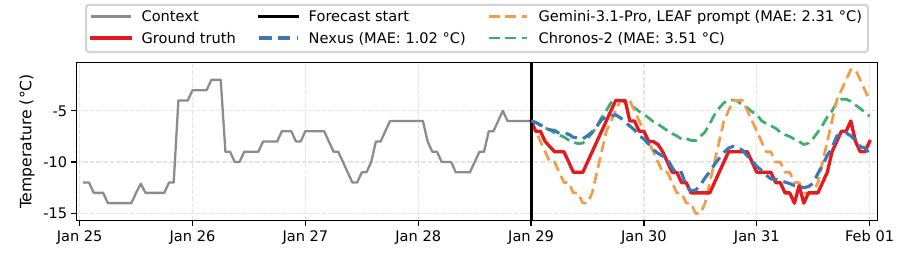}
\caption{Weather, Boston, 72\,h from 2026-01-29}
\end{subfigure}\\[8pt]
\begin{subfigure}{\linewidth}
\centering
\includegraphics[width=0.9\linewidth]{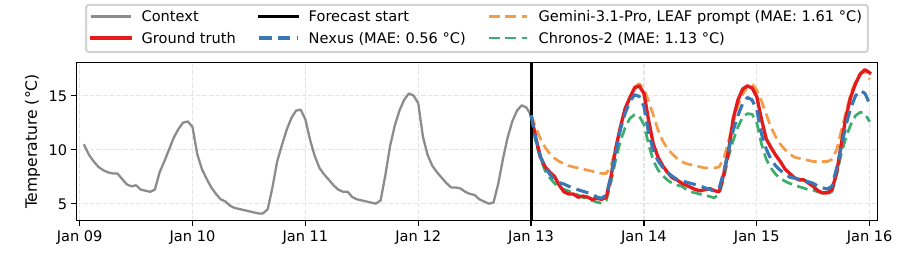}
\caption{Weather, San Francisco, 72\,h from 2026-01-13}
\end{subfigure}
\caption{Qualitative forecast examples on \leaf{}. Each panel shows the context before the forecast start (vertical line), the ground truth, and the forecasts of \ours{}, Gemini-3.1-Pro under the benchmark's prompt with the event record, and Chronos-2. The legend reports each method's error on that window (MAE in $^\circ$C for weather, MAPE otherwise).}
\label{fig:qualitative_leaf}
\end{figure}

\begin{figure}[H]
\ContinuedFloat
\centering
\begin{subfigure}{\linewidth}
\centering
\includegraphics[width=0.9\linewidth]{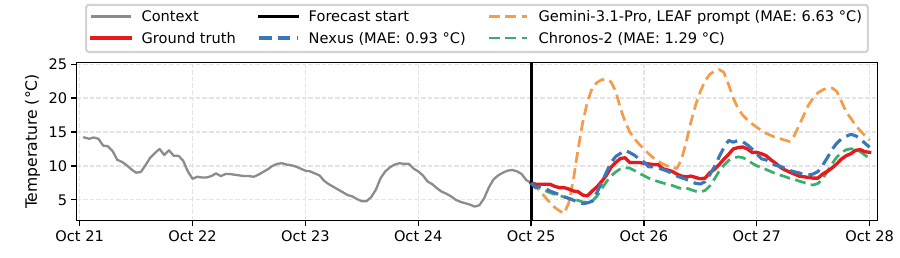}
\caption{Weather, Chicago, 72\,h from 2025-10-25}
\end{subfigure}\\[8pt]
\begin{subfigure}{\linewidth}
\centering
\includegraphics[width=0.9\linewidth]{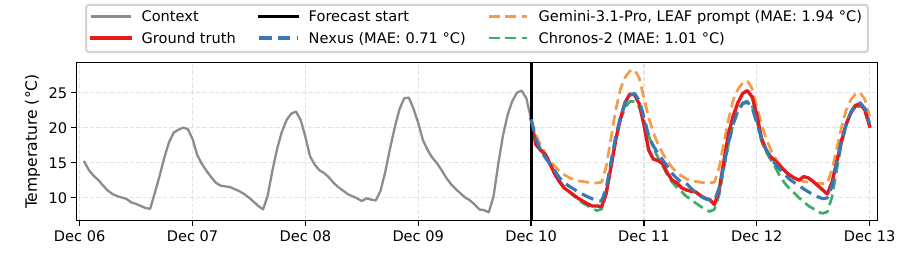}
\caption{Weather, Los Angeles, 72\,h from 2025-12-10}
\end{subfigure}\\[8pt]
\begin{subfigure}{\linewidth}
\centering
\includegraphics[width=0.9\linewidth]{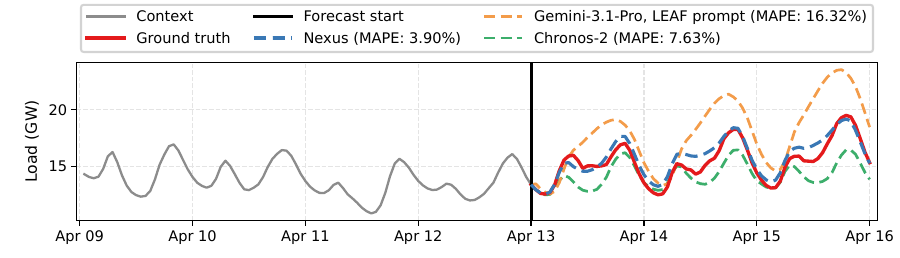}
\caption{Electricity, New York, 72\,h from 2026-04-13}
\end{subfigure}\\[8pt]
\begin{subfigure}{\linewidth}
\centering
\includegraphics[width=0.9\linewidth]{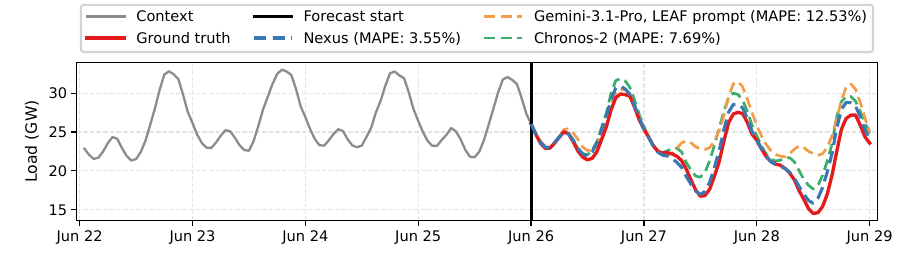}
\caption{Electricity, San Francisco, 72\,h from 2026-06-26}
\end{subfigure}
\caption[]{Qualitative forecast examples on \leaf{} (continued).}
\end{figure}

\begin{figure}[H]
\ContinuedFloat
\centering
\begin{subfigure}{\linewidth}
\centering
\includegraphics[width=0.9\linewidth]{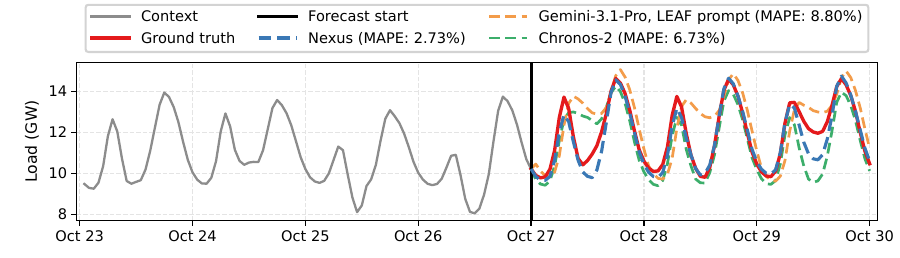}
\caption{Electricity, Boston, 72\,h from 2025-10-27}
\end{subfigure}\\[8pt]
\begin{subfigure}{\linewidth}
\centering
\includegraphics[width=0.9\linewidth]{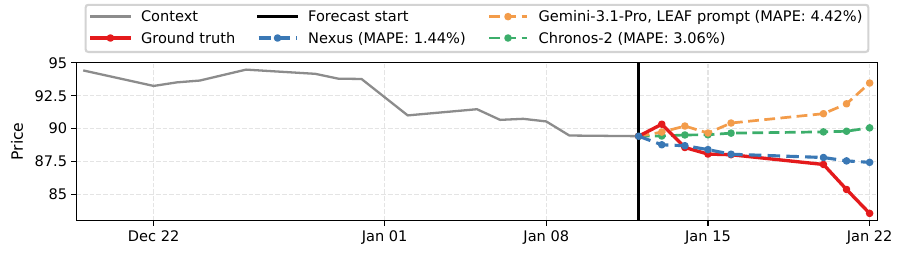}
\caption{Equities, NFLX, 7 trading days from 2026-01-12}
\end{subfigure}\\[8pt]
\begin{subfigure}{\linewidth}
\centering
\includegraphics[width=0.9\linewidth]{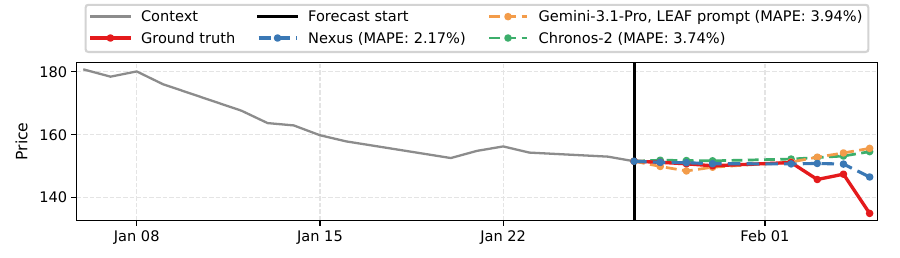}
\caption{Equities, QCOM, 7 trading days from 2026-01-27}
\end{subfigure}\\[8pt]
\begin{subfigure}{\linewidth}
\centering
\includegraphics[width=0.9\linewidth]{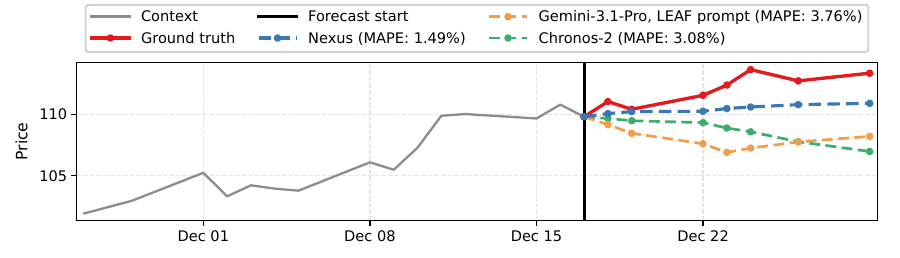}
\caption{Equities, DIS, 7 trading days from 2025-12-17}
\end{subfigure}
\caption[]{Qualitative forecast examples on \leaf{} (continued).}
\end{figure}

\begin{figure}[H]
\ContinuedFloat
\centering
\begin{subfigure}{\linewidth}
\centering
\includegraphics[width=0.9\linewidth]{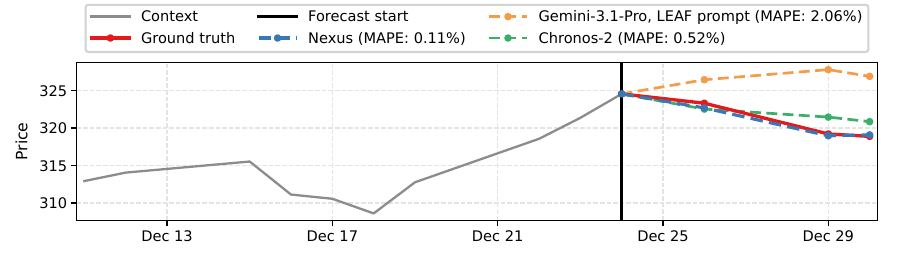}
\caption{Equities, JPM, 3 trading days from 2025-12-24}
\end{subfigure}\\[8pt]
\begin{subfigure}{\linewidth}
\centering
\includegraphics[width=0.9\linewidth]{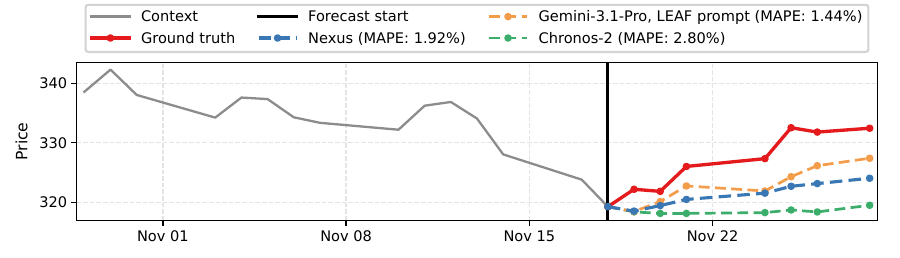}
\caption{Equities, V, 7 trading days from 2025-11-18}
\end{subfigure}\\[8pt]
\begin{subfigure}{\linewidth}
\centering
\includegraphics[width=0.9\linewidth]{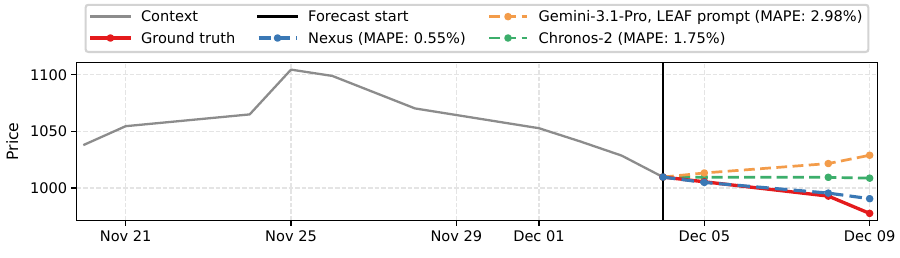}
\caption{Equities, LLY, 3 trading days from 2025-12-04}
\end{subfigure}\\[8pt]
\begin{subfigure}{\linewidth}
\centering
\includegraphics[width=0.9\linewidth]{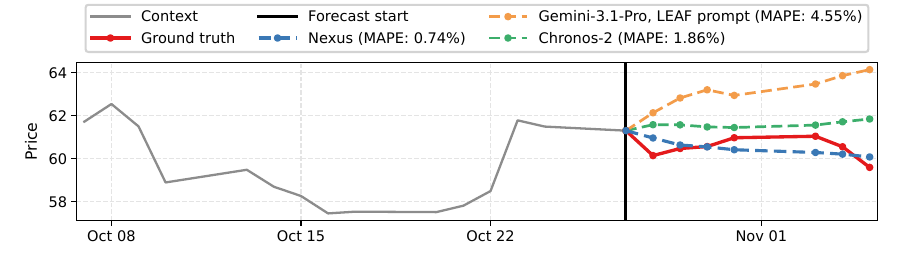}
\caption{Commodities, WTI crude, 7 trading days from 2025-10-27}
\end{subfigure}
\caption[]{Qualitative forecast examples on \leaf{} (continued).}
\end{figure}

\begin{figure}[H]
\ContinuedFloat
\centering
\begin{subfigure}{\linewidth}
\centering
\includegraphics[width=0.9\linewidth]{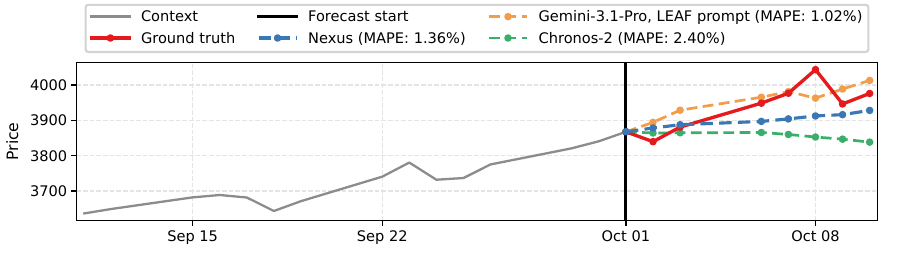}
\caption{Commodities, Gold, 7 trading days from 2025-10-01}
\end{subfigure}\\[8pt]
\begin{subfigure}{\linewidth}
\centering
\includegraphics[width=0.9\linewidth]{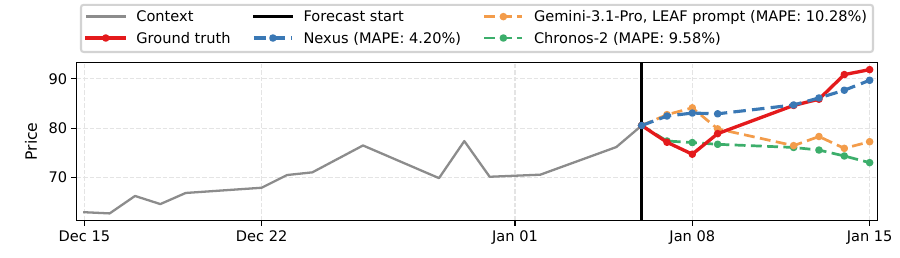}
\caption{Commodities, Silver, 7 trading days from 2026-01-06}
\end{subfigure}\\[8pt]
\begin{subfigure}{\linewidth}
\centering
\includegraphics[width=0.9\linewidth]{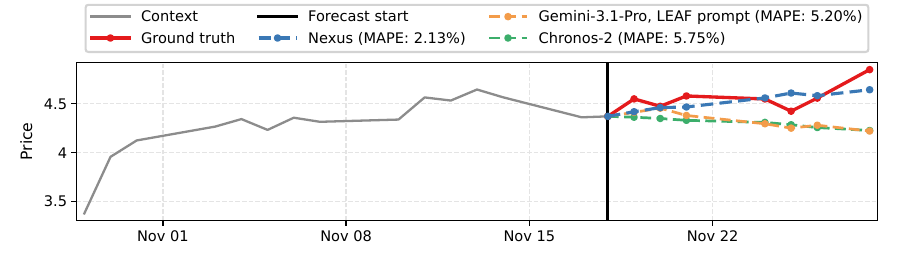}
\caption{Commodities, Natural gas, 7 trading days from 2025-11-18}
\end{subfigure}\\[8pt]
\begin{subfigure}{\linewidth}
\centering
\includegraphics[width=0.9\linewidth]{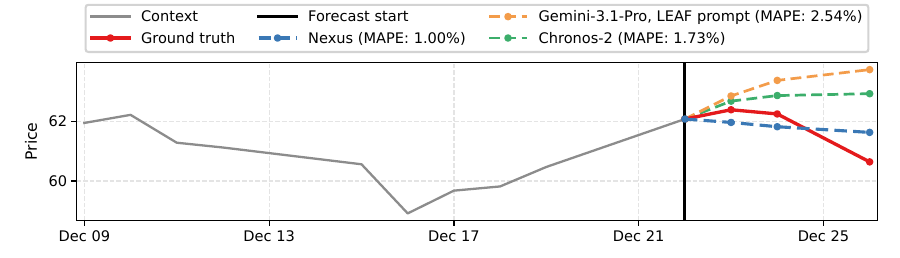}
\caption{Commodities, Brent crude, 3 trading days from 2025-12-22}
\end{subfigure}
\caption[]{Qualitative forecast examples on \leaf{} (continued).}
\end{figure}

\end{document}